# Explainable AI (XAI) for PHM of Industrial Asset: A State-of-The-Art, PRISMA-Compliant Systematic Review

Ahmad Kamal Mohd Nor, Srinivasa Rao Pedapati, Masdi Muhammad,
Universiti Teknologi Petronas, Malaysia.
ahmad_18002773@utp.edu.my

**ABSTRACT**

A state-of-the-art systematic review on XAI applied to Prognostic and Health Management (PHM) of industrial asset is presented. This work provides an overview of the general trend of XAI in PHM, answers the question of accuracy versus explainability, the extent of human involvement, the explanation assessment and uncertainty quantification in PHM-XAI domain. Research articles associated with the subject, from 2015 to 2021 were selected from five known databases following PRISMA guidelines. Data was then extracted from the selected articles and examined. Several findings were synthesized. Firstly, while the discipline is still young, the analysis indicated the growing acceptance of XAI in PHM domain. Secondly, XAI functions as a double edge sword, where it is assimilated as a tool to execute PHM tasks as well as a mean of explanation, particularly in diagnostic and anomaly detection activities, implying a real need for XAI in PHM. Thirdly, the review showed that PHM-XAI papers produce either good or excellent result in general, suggesting that PHM performance is unaffected by XAI. Fourthly, human role, evaluation metrics and uncertainty management are areas requiring further attention by the PHM community. Adequate assessment metrics to cater for PHM need are urgently needed. Finally, most case study featured on the accepted articles are based on real, industrial data, indicating that the available PHM-XAI blends are fit to solve complex, real-world challenges, increasing the confidence in AI model's adoption in the industry.

## 1. INTRODUCTION

### 1.1. General Progress in AI

Artificial Intelligence (AI) continues its extensive penetration into emerging markets, driven by the untapped opportunities of the 21$^{st}$ century and backed by steady and sizeable investments. In the last few years, AI-based research shows much concentration in areas such as large-scale machine learning, deep learning, reinforcement learning, robotic, computer vision, natural language processing and internet of Thing (IoT) [1]. According to the first AI experts report in the *One Hundred Year Study on Artificial Intelligence,* AI ability will be heavily embodied in transportation, home robotics, healthcare, education, security and safety as well as entertainment in north American cities by the 2030's [1]. The increasing data volume, breakthrough in machine learning, coupled with the pressing need to be more efficient and innovative democratize AI to the global scene. A survey done by McKinsey recorded an annual increase of 30% in AI investment from the 2010 to 2013 and 40% from the 2013 to 2016. In 2016 alone, the total global investment amounted to 26 to 39 billion dollars by tech firms and external investments [2]. By the 2030, AI could potentially value up to 15 trillion dollars in global GDP growth thanks to automation and product innovation while reducing approximately 7 trillion dollars in operational cost [3]. AI-driven technology will lead to incremental change in labour market requirement, where increasing technological ability, together with higher cognitive and social-emotional skills are needed to support AI-based infrastructures, whereas manual and basic cognitive skills will experience lesser demand [4].

AI is a technical discipline defined as the science to make computers do things that would require intelligence if done by humans [5]. The reasoning of AI imitates natural law translated in working algorithms [6]**.** Some important fields in AI research include Expert System (ES), consisting of rule-based reasoning (RBR), case-based reasoning (CBR), and Fuzzy Systems (FL) along with Machine Learning (ML) models such as Neural Network (NN), Support Vectors, Deep Learning (DL) and Heuristic Algorithms [7]. The availability of parallel Graphics Processing Unit (GPU) and open-source development tools open the door for literally everyone to solve technical challenges, sometimes surpassing human performance [8, 9]. These abilities and specialized tools make AI so appealing technically infused domains such as healthcare [5], computer vision [8], image processing [6] and reliability engineering [7].

### 1.2. AI in PHM

Machine learning in general, and more specifically deep learning, has been part of the reliability research landscape including PHM [10,11,12]. PHM provides guidelines and frameworks to safeguard the healthy state of assets. It minimizes risks, maintenance cost and workload, thus optimizing maintenance activities. PHM is defined in IEEE standard as "a maintenance and asset management approach utilizing signals, measurements, models, and algorithms to detect, assess, and track degraded health, and to predict failure progression" [13]. Accordingly, 3 types of PHM activities are distinguished: failure prognostic, diagnostic, and anomaly detection. Prognostic is the action of determining the Remaining Useful Life (RUL) or the left-over operational time of an asset before failure [12]. Diagnostic is the action of classifying failure and, to some extent, discovering the detailed root cause of the failure while anomaly detection consists of identifying unusual patterns going against the normal behaviour of operational indicators [14,15].

Various literatures support the idea of AI as being at the forefront in PHM [10,15]. To mention a few: Long Short-Term Memory (LSTM) neural network is employed in [16] with degradation image to estimate the RUL of rotating machinery. A regression tree is used to predict the RUL of central heating and cooling plant (CHCP) equipment in [17]. In [18], the combination of Logistic Regression (LR) with L2 Support Vector Machine (SVM) are proposed for gas circulator unit prognostic. Random Forest (RF) is employed to diagnose fault for semiconductor equipment failure in [19]. Convolutional and fully connected layers with Softmax activation are proposed in [20] to diagnose rotating machine issues. Gradient Boosted Decision Tree (GBDT) outperformed other methods in anomaly detection of hard drives in [21].

### 1.3. Black Box AI Problem

Though very powerful, many AI methods are black box in nature, meaning that the inner mechanism to produce output in these techniques are not known [22,23]. Obviously, this opacity is an obstacle in AI penetration across many sensitive or high-stake areas such as medical, banking, finance, defence, or the even the common industry [24,25]. The end user and experts of the domain in question need the assurance that the model's inner process is understandable [26]. The opaqueness adds operational and confidentiality hazards, bias, or non-ethical outputs risks [27]. The non-transparency discourage a responsible exploitation of AI decisions [28], model troubleshooting [29] and improvement [26]. Moreover, it further complicates the question of responsibility ownership in the case of wrong decision [30]. Finally, with the increasing scrutiny and regulation on AI usage, the need to make AI methods as transparent as possible is pressing. This includes the General Data Protection Regulation (GDPR) in the European Union (EU) and the Ethics Guidelines for Trustworthy Artificial Intelligence presented by the European Commission High-Level Expert Group on AI [31,32,33] .

### 1.4. The Need for XAI

XAI is a discipline dedicated in making AI methods more transparent, explainable, and understandable to end user, stakeholders, non-experts, and non-stakeholders alike to nurture trust in AI. The growing curiosity in XAI is mirrored by the spike of interest in the search term "Explainable AI" since the 2016 and the rising number of publications throughout the years [32]. DARPA developed the *Explainable AI (XAI) Program* in 2017 while the Chinese government announced *The Development Plan for New Generation of Artificial Intelligence* in the same year, both promoting the dissemination of XAI [34].

The general needs for XAI are as follows:

1. Justify model’s decision, identifying issues, and enhancing AI models.
2. Obey AI regulation and guidelines in usage, bias, ethical, dependability, accountability, safety, and security.
3. Allow user to confirm the model's desirable features, promote engagement, obtain fresh insights into the model or data, and augment human intuition.
4. Allow user to better optimise and focus their activities, efforts, and resources.
5. Support model development when it is not yet considered as reliable.
6. Encourage the cooperation between AI experts and external parties.

### 1.5. Review Motivation

The objective of this review is to present an overview of XAI applications in PHM of industrial asset. Preferred Reporting Items for Systematic Reviews and Meta-analyses (PRISMA) guidelines was employed [35]. It is an evidence-based guidelines that ensure comprehensiveness, reducing bias, increasing reliability, transparency, and clarity of the review with minimum item [36,37]. PRISMA is 27-checklist guidelines that need to be satisfied as best as possible for best practice in systematic review redaction. However, items 12, 13e, 13f, 14, 15, 18-22 and 24 were omitted as they were not treated this systematic review.

Rationalities motivating the compilation of this review:

1.a. ***Global interest in XAI:*** According to survey, the general curiosity toward XAI has surged since 2016 [9]. **Figure 1** show the search interest expressed for “Explainable AI” term in Google search, with 100 as being the peak popularity for any term.

1.b. ***Specialized Reviews:*** In the early years of XAI, several general surveys on XAI methods were written [32,34]. More recently, as the discipline grows, more specialized works emerged. [24], [27] and [30] present reviews on XAI in healthcare, [22] in pathology, [23] in psychology, [29] in fintech management, [33] in neurorobotics, [25] in drug discovery, and [31] in plant biology. It is thus urgent to produce an analytical compilation of PHM-XAI works which is still absent.

1.c. ***PHM Nature & Regulation:*** PHM is naturally related to high investment and safety sensitive industrial domains. Moreover, it is pressing to ensure the use of well-regulated

AI in PHM. It is thus obvious for XAI to be promoted as much as possible and its know-how disseminated for the benefit of PHM actors.

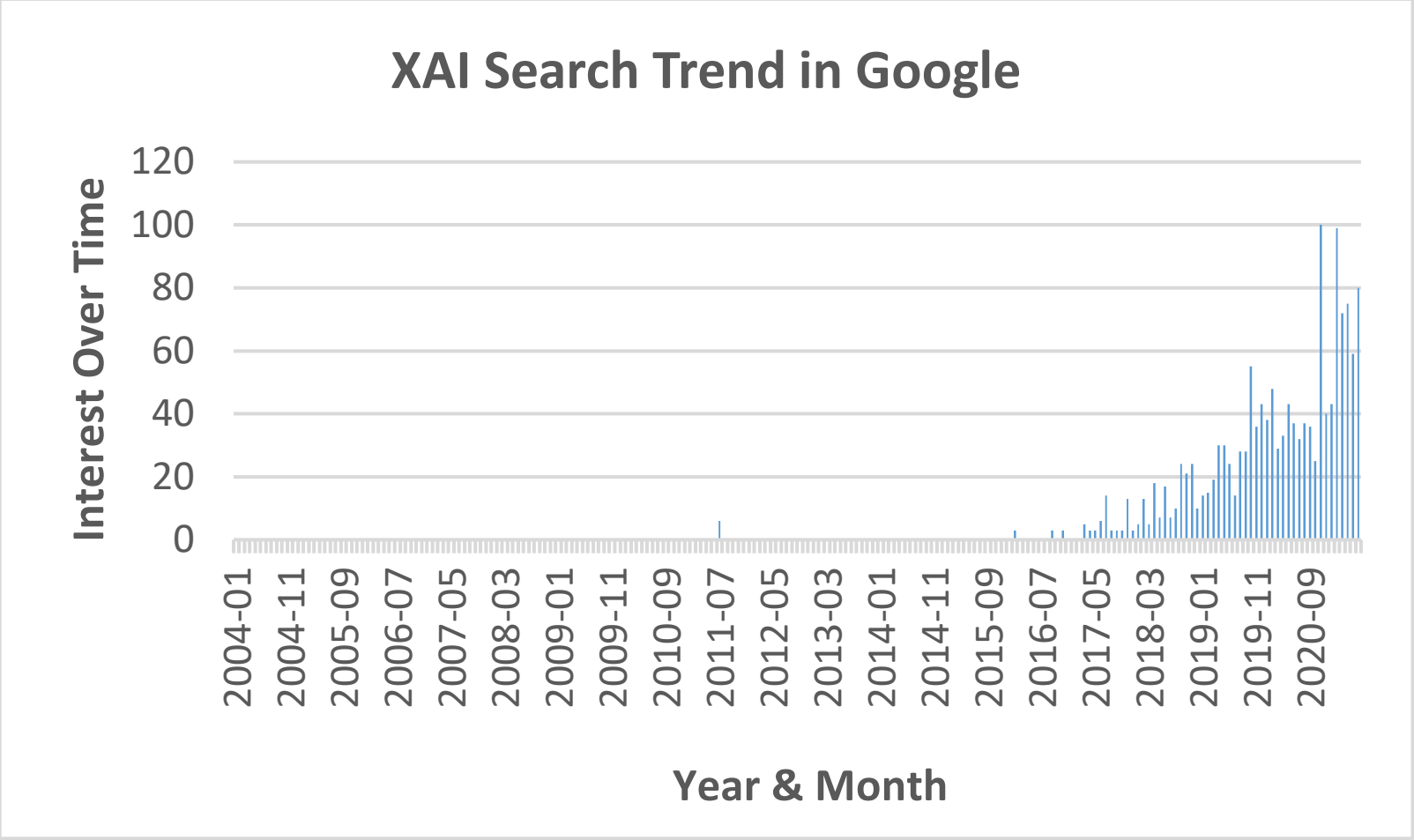

**Figure 1.** Interest Shown for *Explainable AI* Term in Google Search

The review goals are achieved by addressing the following points:

1.d. General trend: Interest in XAI, overview of the XAI approach employed, the repartition of the said methods according to PHM activities and the type of case study involved.

1.e. Accuracy versus the explainability power: According to DARPA, model's accuracy performance is inversed to its explainability prowess [34].

1.f. XAI role in assisting or burdening PHM tasks.

1.g. The challenges in PHM-XAI progress: Cross checks were done with the general challenges raised in [9,26,28,32], specifically:

   I. The lack of explanation evaluation metrics.
   II. The absence of human involvement for enhancing the explanation effectivity.
   III. The omission of uncertainty management in the studied literature.

This paper is organized as follow: In Section 2, the methodology is introduced, followed by the results presentation in Section 4. Then, the discussion is elaborated in Section 5. Finally, concluding remarks are presented in last section.

## 2. METHOD

A single reviewer was associated with the search, screening, and data extraction of the articles. Thus, no disagreement occurred in all the steps mentioned. Only peer reviewed journal articles on PHM-XAI of industrial asset between 2015 to 2021 in English language were selected. Five publication databases consisting of ScienceDirect (until 17/02/21), IEEE Xplore (until 18/02/21), SpringerLink (until 22/02/21), ACM Digital Library (until 28/05/21) and Scopus (until 27/02/21) were explored. Advance search was used but since each database features are different, specific strategy was adopted. In IEEE Xplore, search was done in the *Abstract* and *Document title* fields only as these are the most relevant options. The database also authorises search within the obtained results in the *Search within results* field. Wildcard was not used in IEEE Xplore even though permitted. Comprehensive search in the title, abstract and keywords fields were done in ScienceDirect and Scopus: *Title, abstract and author-specified keywords* fields for ScienceDirect and *Search within Article title, Abstract, Keywords* fields for Scopus. However, contrary to Scopus, ScienceDirect does not support wildcard search, thus wildcard was only used in Scopus. In SpringerLink, the *with all the words* field was used altogether with wildcards. In ACM, both the ACM Full-Text collection and ACM Guide for Computing Literature were explored. *Search Within* option in the title, abstract and keywords was executed with wildcard. Once done, the screening of duplications was performed using Zotero software. The full research strategy is listed in **Appendix A**.

Below screening steps were executed one after another for the rest of the result, with each screening step starting in the title, then the abstract and next, the keywords:

2.a. Verify if article type is research article.

2.b. Exclude non PHM articles by identifying absence of commonly employed PHM terms such as *prognostic*, *prognosis*, *remaining useful life*, *RUL*, *diagnostic*, *diagnosis*, *anomaly detection*, *failure*, *fault*, or *degradation*.

2.c. Exclude non XAI articles by identifying absence of commonly used XAI terms which are *explainable*, *interpretable* and *XAI*.

2.d. Exclude non PHM-XAI articles by identifying the absence of both PHM and XAI terms as respectively indicated in step 2.b. and 2.c.

2.e. Exclude articles related to medical applications or network security.

Finally, the context of the articles was examined on the remaining works for final screening to retain only the desired articles.

The data extracted from the articles was gathered in an MS. Excel file. Directly retained variables were author, publication year, title, publisher, and publication/journal name. Once final selection was achieved, the article context analysis (full text body check) was performed, and further information was extracted:

2.f. PHM activity category – Either anomaly detection, prognostic, or diagnostic. Structural damage detection as well as binary failure prediction were considered as diagnostic.

2.g. XAI approach employed – The category of XAI method.

2.h. Recorded performance – Reported accuracy. Some papers clearly claim the comparability or the superiority of the proposed method over other tested methods. In the case where comparison is not done, the reported standalone accuracy result was evaluated and classified as either “good” or “satisfying” for a medium outcome or deemed “very good” or “excellent” if the accuracy is near perfect. When a method is superior to the rest, it was classified as “very good” unless detailed as only “good”. When mixed performance of good and very good are recorded for the same method, it was quantified as only “good”.

2.i. XAI role in assisting PHM task – Here, XAI role in strengthening PHM ability was investigated.

2.j. The existence of explanation evaluation metrics.

2.k. Human role in PHM-XAI works.

2.l. Uncertainty management – Uncertainty management in either any of the stage of the PHM method or XAI approach increases the possibility for adoption by user due to additional surety.

2.m. Case study type – real or simulated. Real was considered when data is coming from a real mechanical device. Simulated was considered when data is generated using simulation software.

The outputs were presented in either pie chart or column graph for comparison.

2.n. Pie chart – PHM activity category, explanation metric, human role, uncertainty management.

2.o. Column graph – PHM-XAI yearly trend, XAI approach employed, recorded performance, XAI role in assisting PHM task.

## 3. RESULTS

### 3.1. Screening Results

3048 papers were selected from the databases according to the applied keywords with their respective quantity as shown in **Table 3**. 288 articles were screened out as duplicates. Out of the 2760 remaining, 25 papers had to be screened out as they are editorial or news documents. Then, 70 papers were selected according to criteria 2.a. - 2.e. from the remaining 2735 articles. Finally, only 35 papers were selected as 35 articles were deemed not relevant with the review topic after context verification. The final selection and the excluded articles with their decision motives are disclosed respectively in **Table 1** and **Table 2**. The PRISMA flow diagram of the selection and screening process is disclosed in **Figure 2**.

The repartition of the selected articles' PHM domain as well as its publisher are presented in **Figure 3** and **Figure 4**. The repartition of the excluded articles' PHM domain as well as its publisher are presented in **Figure 5** and **Figure 6**. As can be seen from **Figure 3**, diagnostic research holds the most share in PHM-XAI articles. **Figure 4** illustrates that IEEE and Elsevier as being the biggest source of the accepted articles.

Numerous unselected publications, though related to XAI, correspond to process monitoring research as shown in **Figure 5**. These works were excluded as they are closely related to quality context rather than failure. Some are focused on product instead of the industrial asset. Furthermore, the anomaly described is seldom associated with process disturbance rather than failure degradation. Studies concerning network security were also omitted. Here too, most of the excluded papers come from Elsevier and IEEE as confirmed by **Figure 6**, further showing that these publishers are the main sources of much XAI related articles.

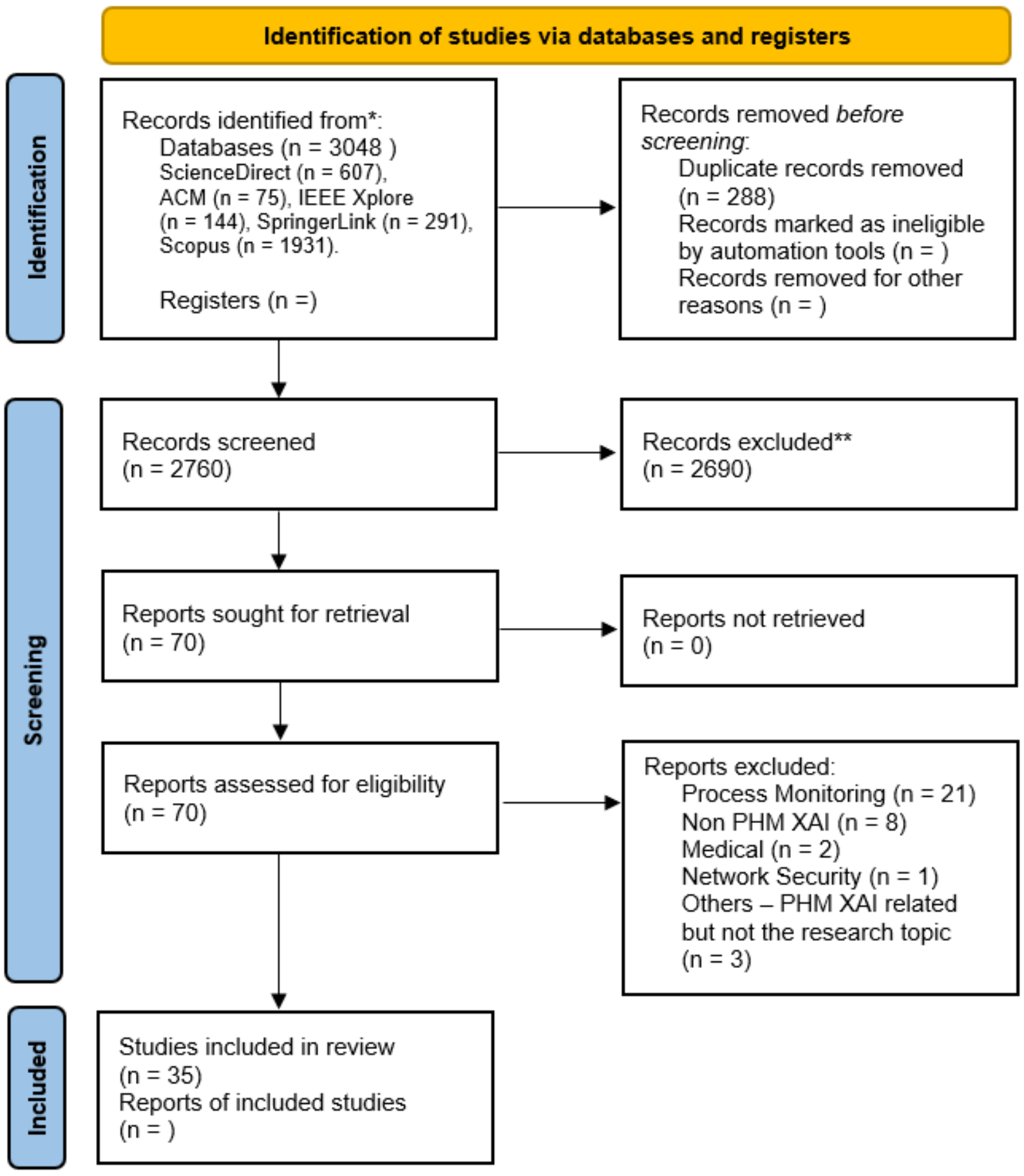

**Figure 2.** PRISMA Flow Diagram Search Strategy

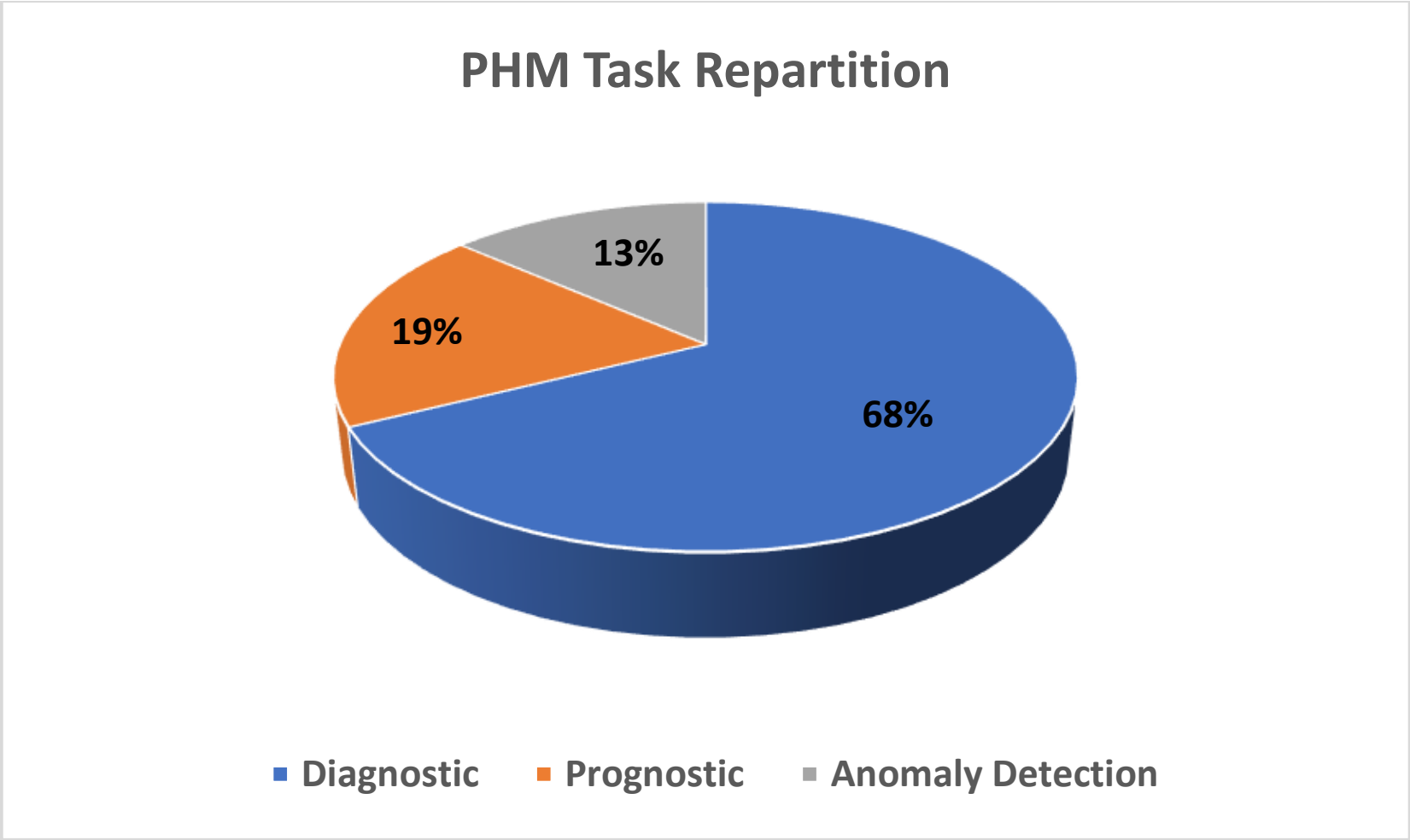

**Figure 3.** PHM Tasks of Accepted Article

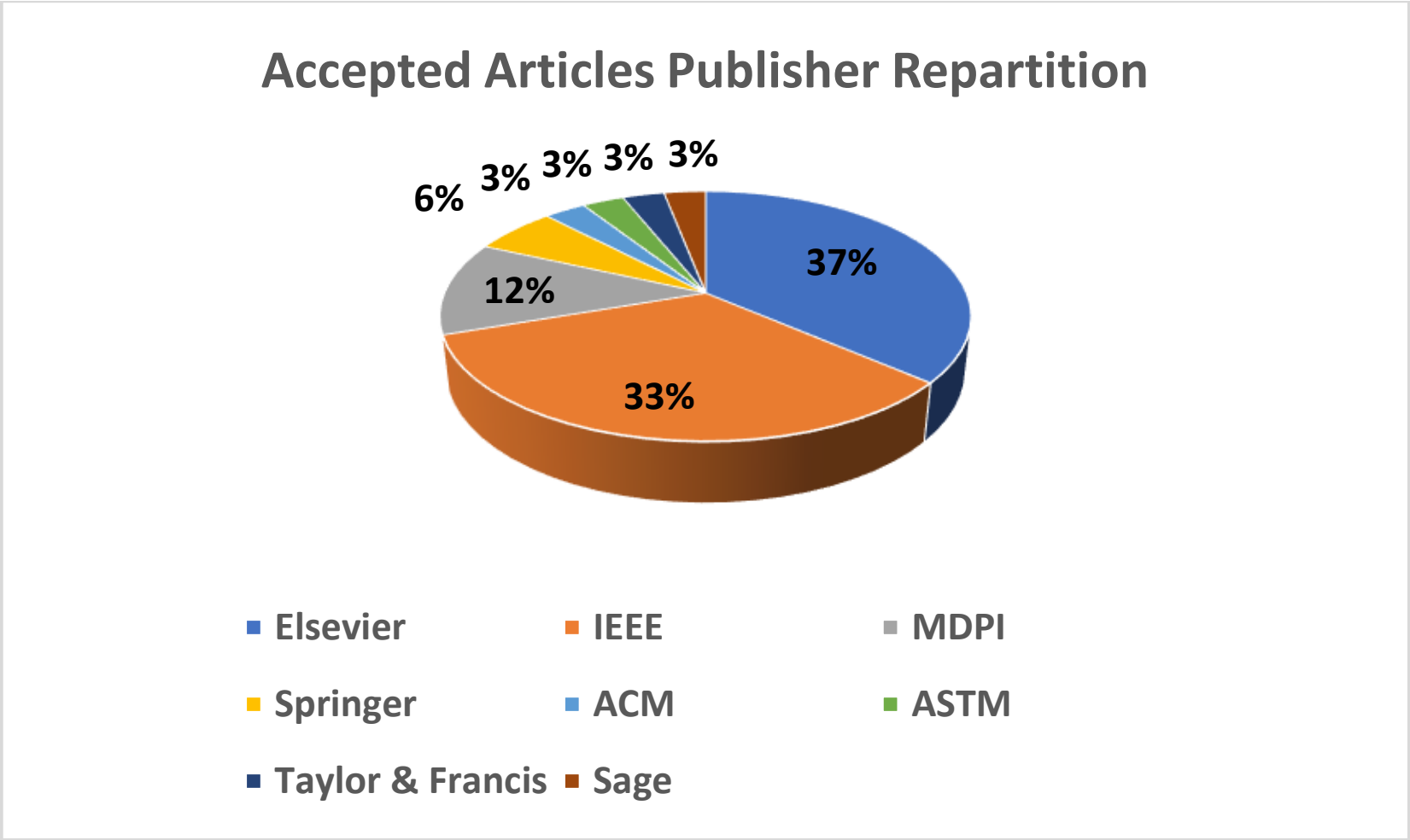

**Figure 4.** Accepted Articles Publisher Repartition

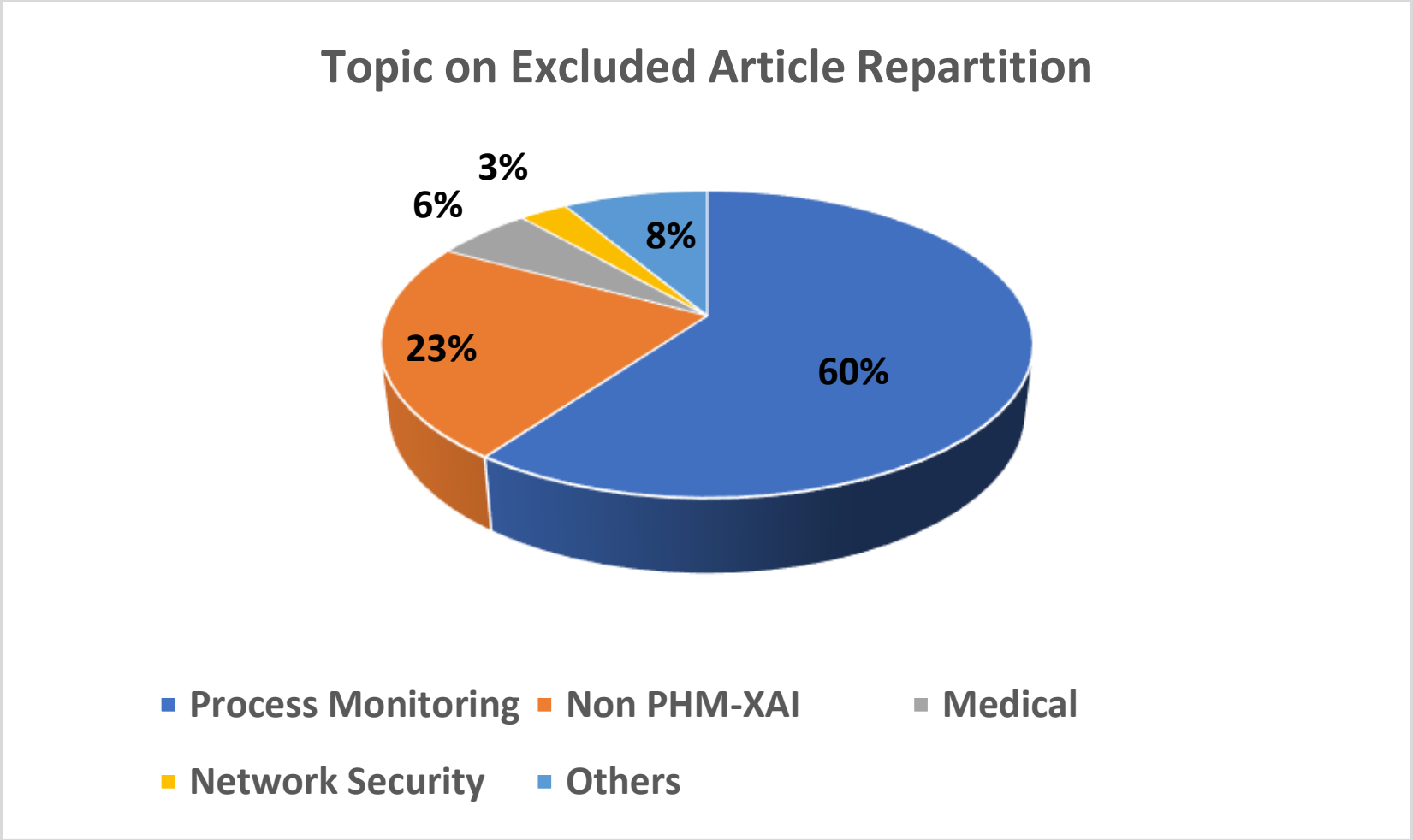

**Figure 5.** Excluded Articles Topic Repartition

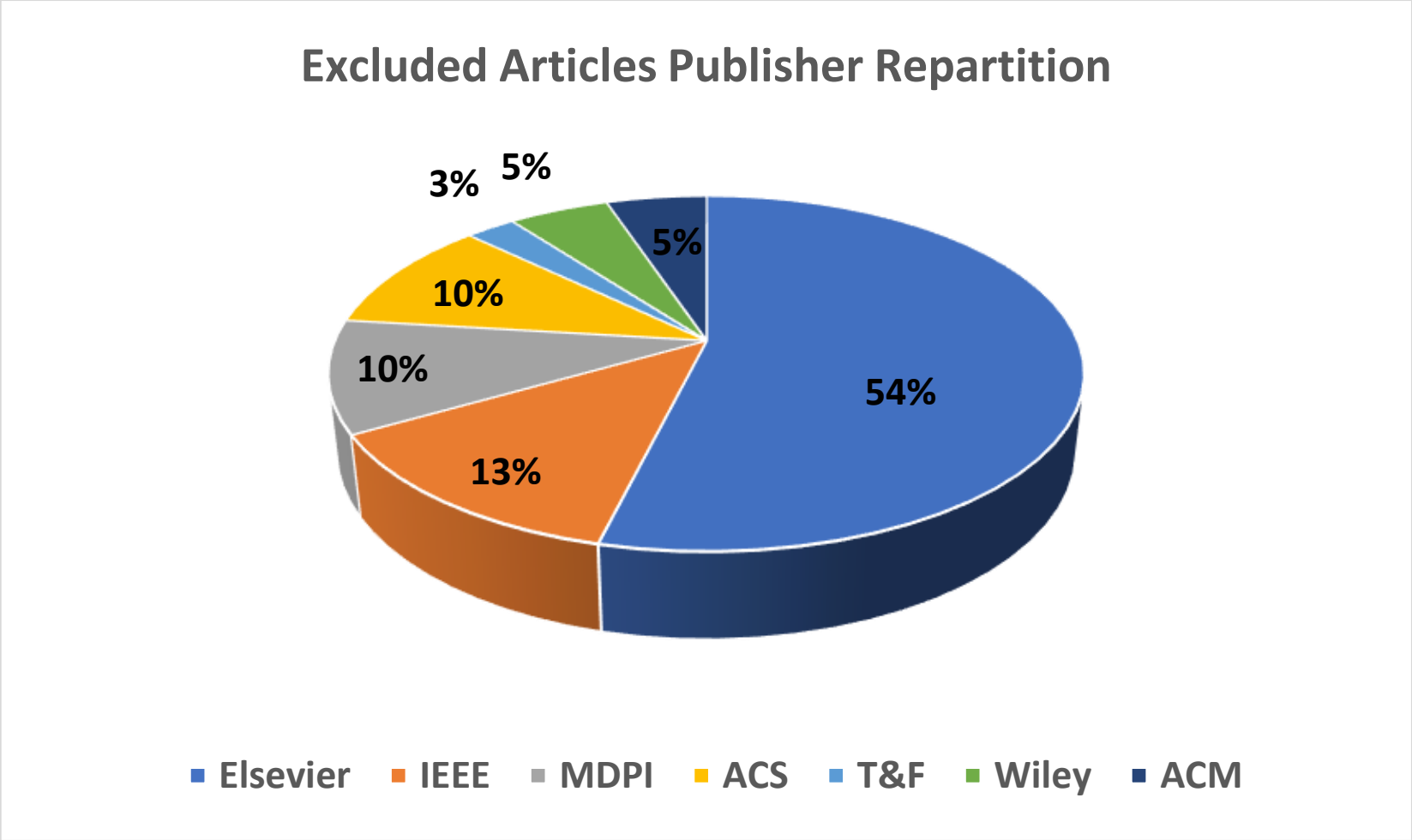

**Figure 6.** Excluded Articles Publisher Repartition

## 4. DISCUSSION

### 4.1. General Trend in PHM-XAI

As shown in **Table 1** and summarized in **Figure 7**, accepted articles according to publication year show an upward trend, with a major spike in 2020, indicating a growing interest in XAI from the PHM researchers. The number of accepted articles is still very small, reflecting infancy state of XAI in PHM. Diagnostic domain occupies the majority share amongst the accepted works as presented in **Figure 3**. Looking at the *XAI Assist PHM* column in **Table 1**, it can be deduced that XAI boosts diagnostic ability. Drawing a parallel between the two information from **Figure 3** and **Table 1**, it can be inferred that XAI is particularly appealing to diagnostic as it can be applied directly as a diagnostic tool or in addition to other methods. XAI could provide additional incentive to diagnostic whose main objective is to discover the features responsible for the failure as shown in **Figure 8**. This interesting point signifies that the diagnostic task in these papers are dependent on XAI. Therefore, XAI is not only a supplementary feature in diagnostic but also an indispensable tool. The same phenomenon is observed in anomaly detection as presented in **Figure 8**. Knowing the cause of anomaly could potentially avoid false alarm, preventing resource wastage. XAI can thus be employed as a double-edged sword to execute PHM tasks and explains it. Additionally, interpretable models, rule & knowledge-based models as well as attention mechanism are the three most employed approaches as illustrated in **Figure 9**.

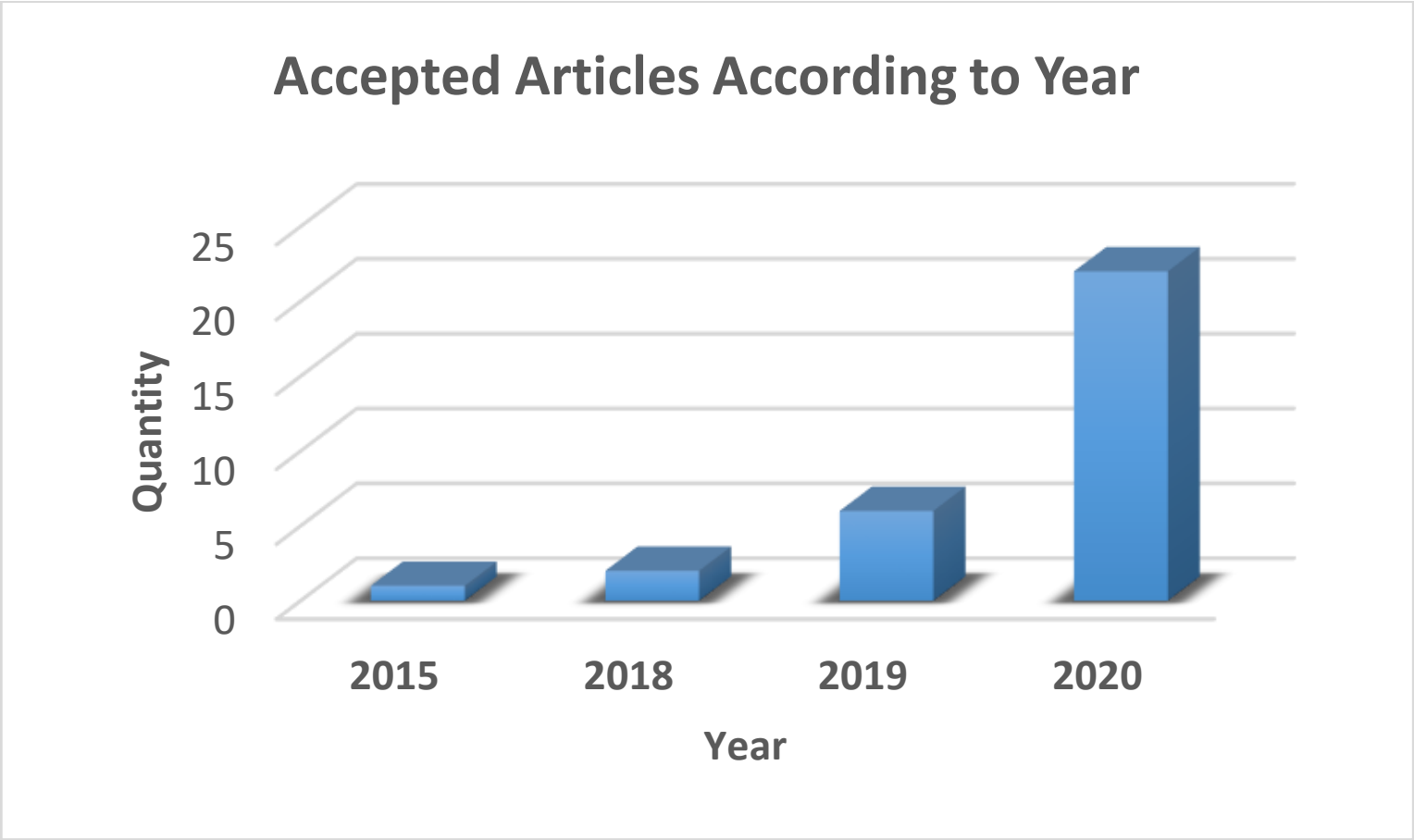

**Figure 7.** Accepted Articles According to Year

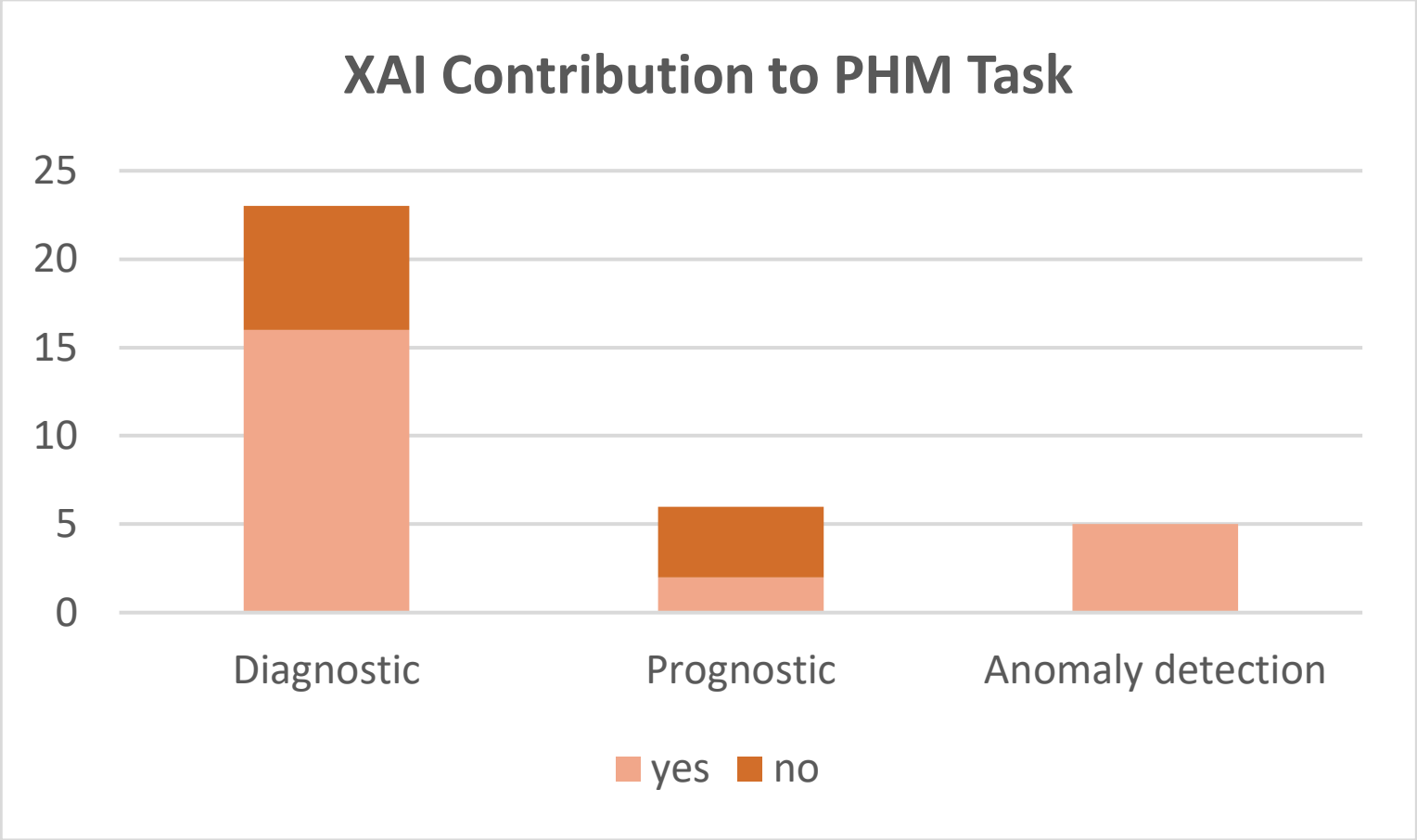

**Figure 8**. XAI Assistance to PHM Tasks

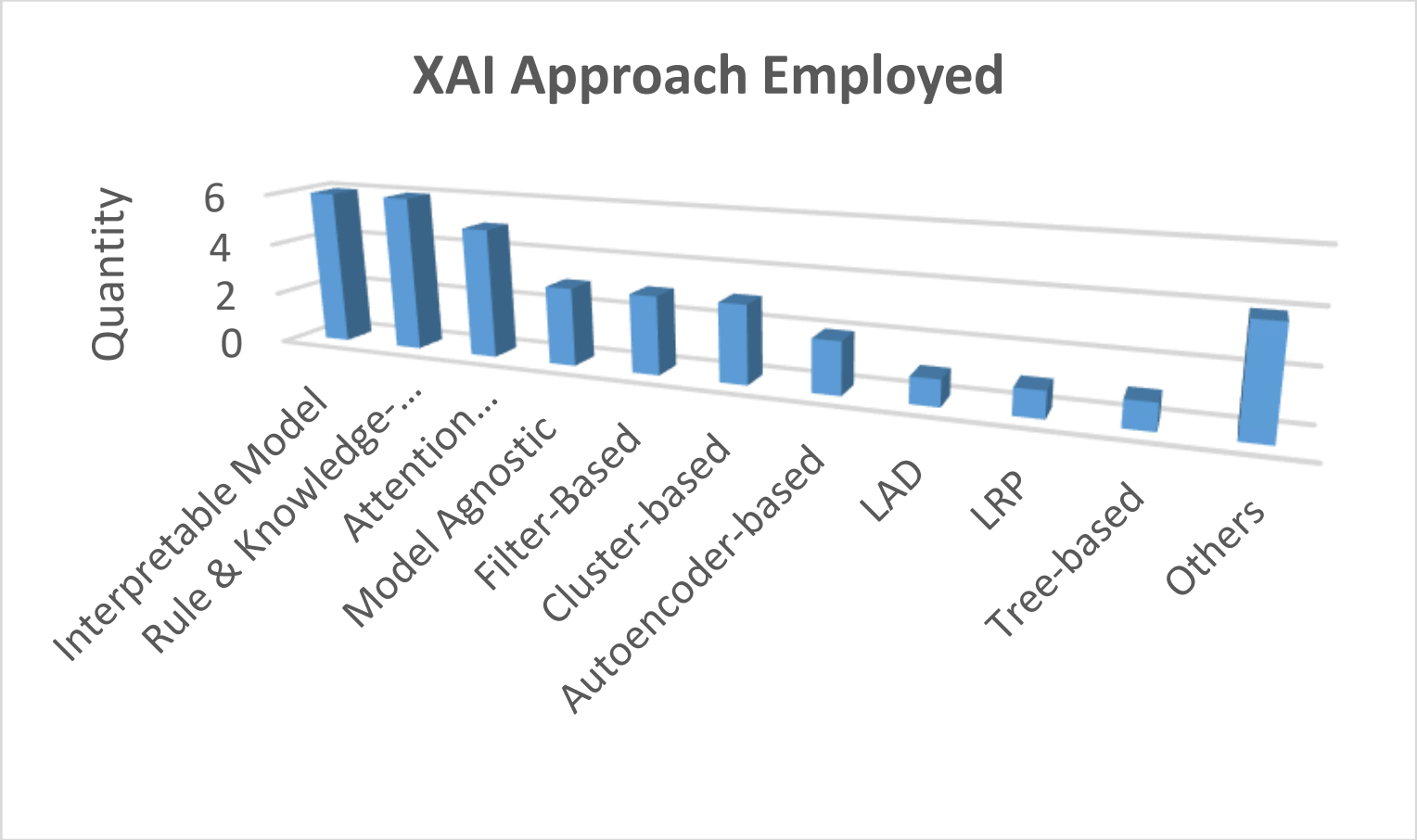

**Figure 9.** XAI Approach Type in Selected Articles

4.2. Accuracy vs Explainability

**Table 1** reveals that some XAI approaches that directly assist PHM tasks achieved excellent performance. Additionally, the recorded PHM performance of both XAI and non XAI methods (works that depend on XAI for explanation only) are mostly very good for diagnostic and prognostic, as depicted in **Figure 10**. In brief, no bad result was recorded as confirmed by **Figure 10**. Whether the results are contributed by XAI or not, it can safely be concluded that explainability does not affect the tasks' accuracy in the studied works.

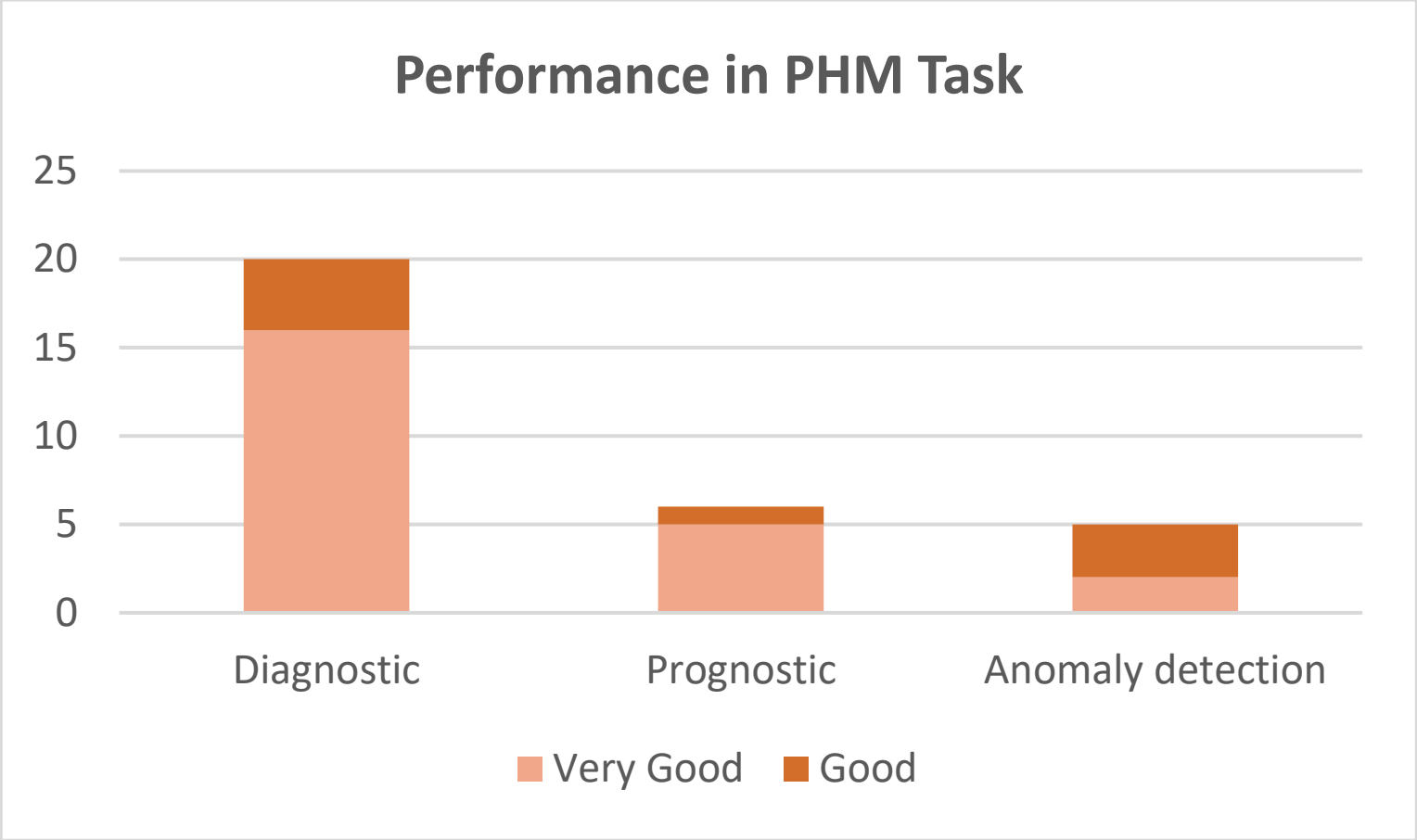

**Figure 10.** Performance of AI Models According to Task

4.3. Human Role in XAI & Explainability Metrics

Very little role was played by human in the examined works as illustrated in **Figure 11** Human participation is vital for evaluating the generated explanation. Human involvement is encouraged for the development of interactive AI, where expert's opinion presents an additional guarantee in AI performance. In a more serious observation, the usage of explanation evaluation metrics is nearly non-existent as presented in **Figure 12**. These measures could help researchers and developers to evaluate the explanation quality. It is recommended that adequate assessment method for PHM, considering security and safety risk, maintenance cost, time, and gain to be developed and adopted. [73] presents an overview of explanation metrics and methods. [74] studies the effectiveness of explanation from experts to non-experts while [75] proposes a metric to assess the quality of medical explanation.

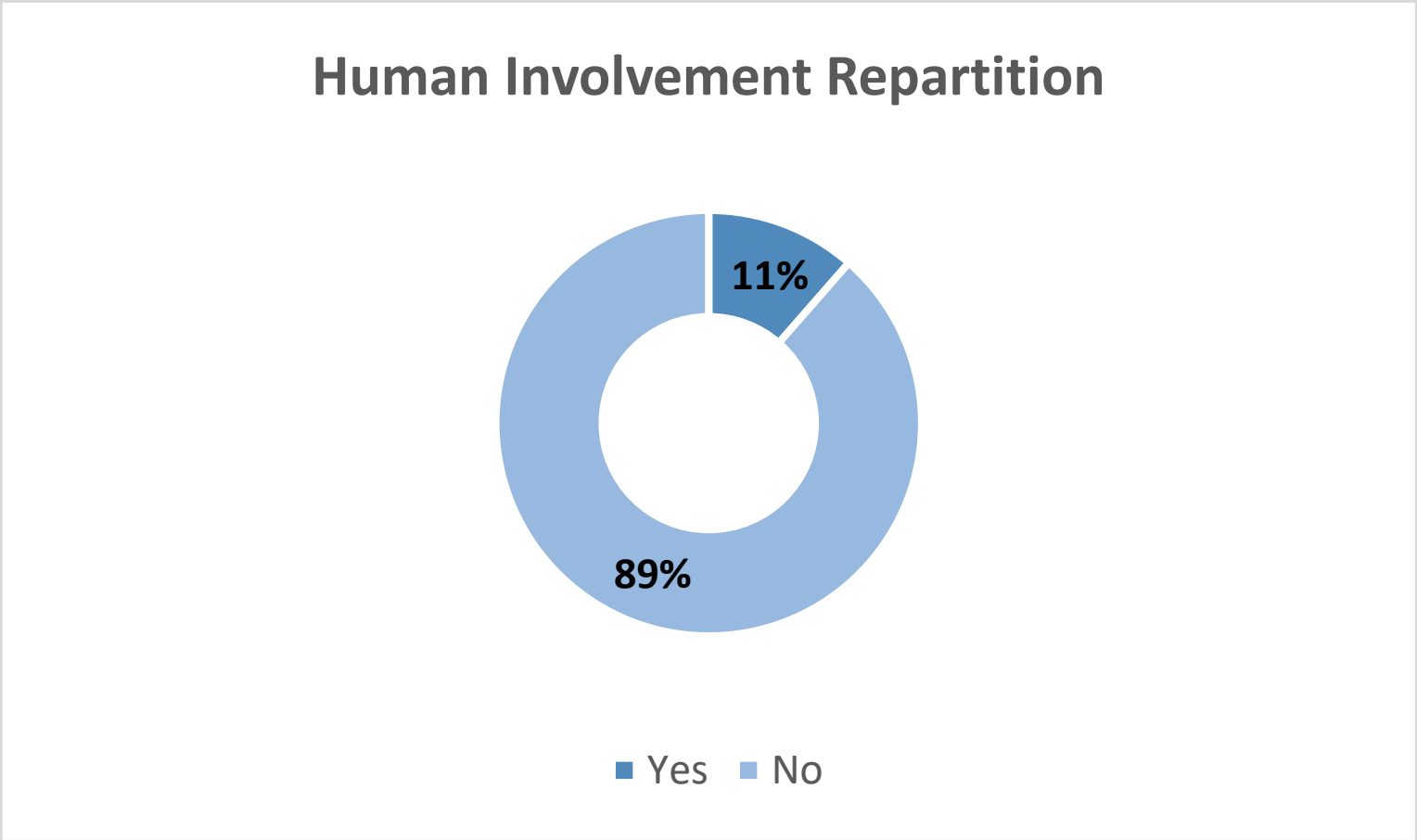

**Figure 11.** Human Involvement in Selected Works

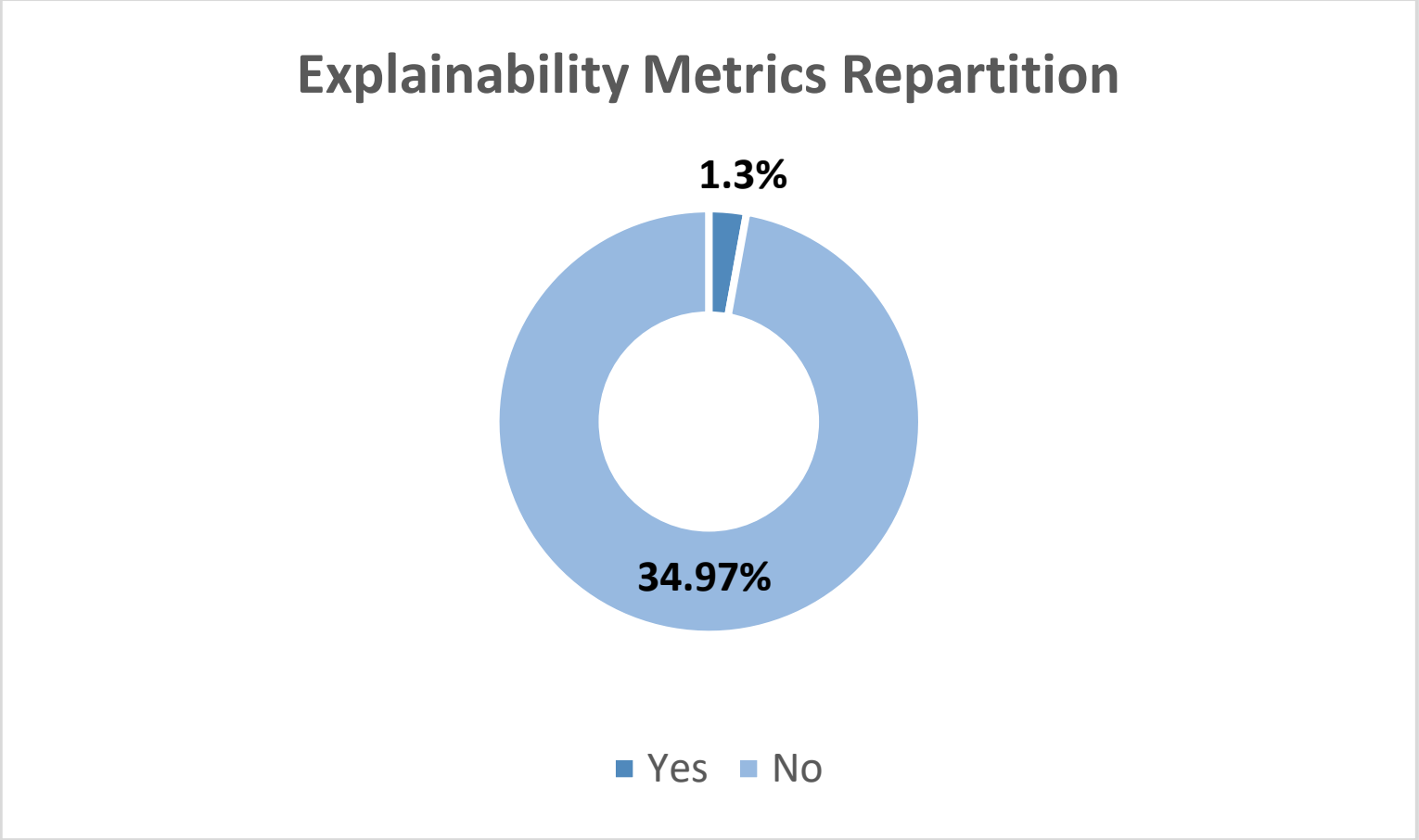

**Figure 12**. Inclusion of Explainability Metrics in Selected Works

4.4. Uncertainty Management

Various types of uncertainty management approach are adopted in different stages in the studied works as detailed in **Table 1**. Uncertainty management gives additional surety to users to adopt PHM-XAI methods compared to point estimate models. Furthermore, uncertainty quantification is vital to provide additional security to AI infrastructure against adversarial example, either unintentionally or motivated by attack. This measure could minimize the risk of wrong explanation being produced from unseen data due to adversarial example. It is observed in **Figure 13** however, that much improvement is still required in this area.

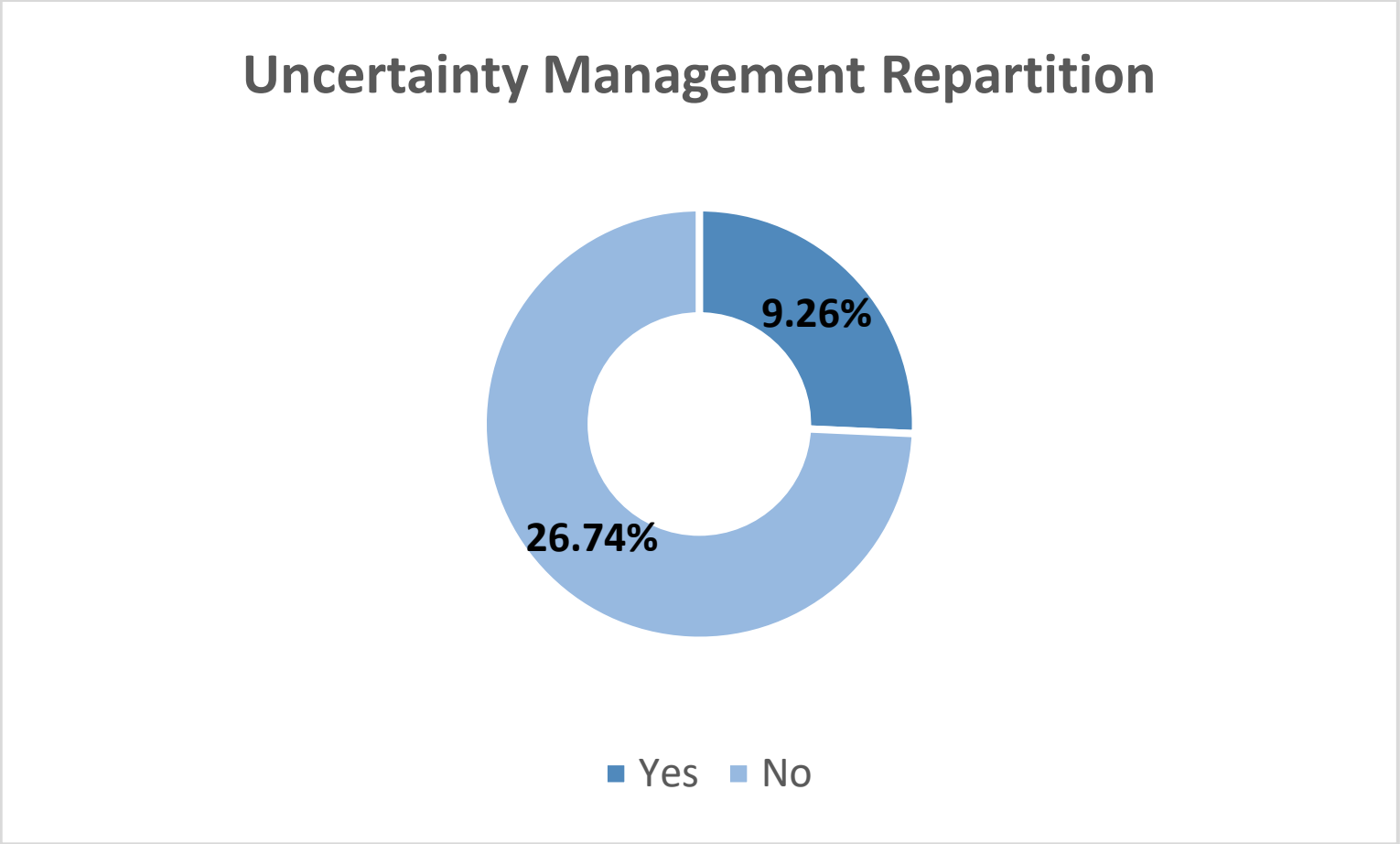

**Figure 13.** Uncertainty Management Inclusion in Selected Works

4.5. Case Study Type

Real industrial data is mostly used in the case studies to demonstrate the effectiveness of XAI as reflected in **Figure 14**. This positive outlook proves that the available PHM-XAI combinations are able to solve real world industrial challenges with at least a good performance, boosting the confidence in AI model's adoption.

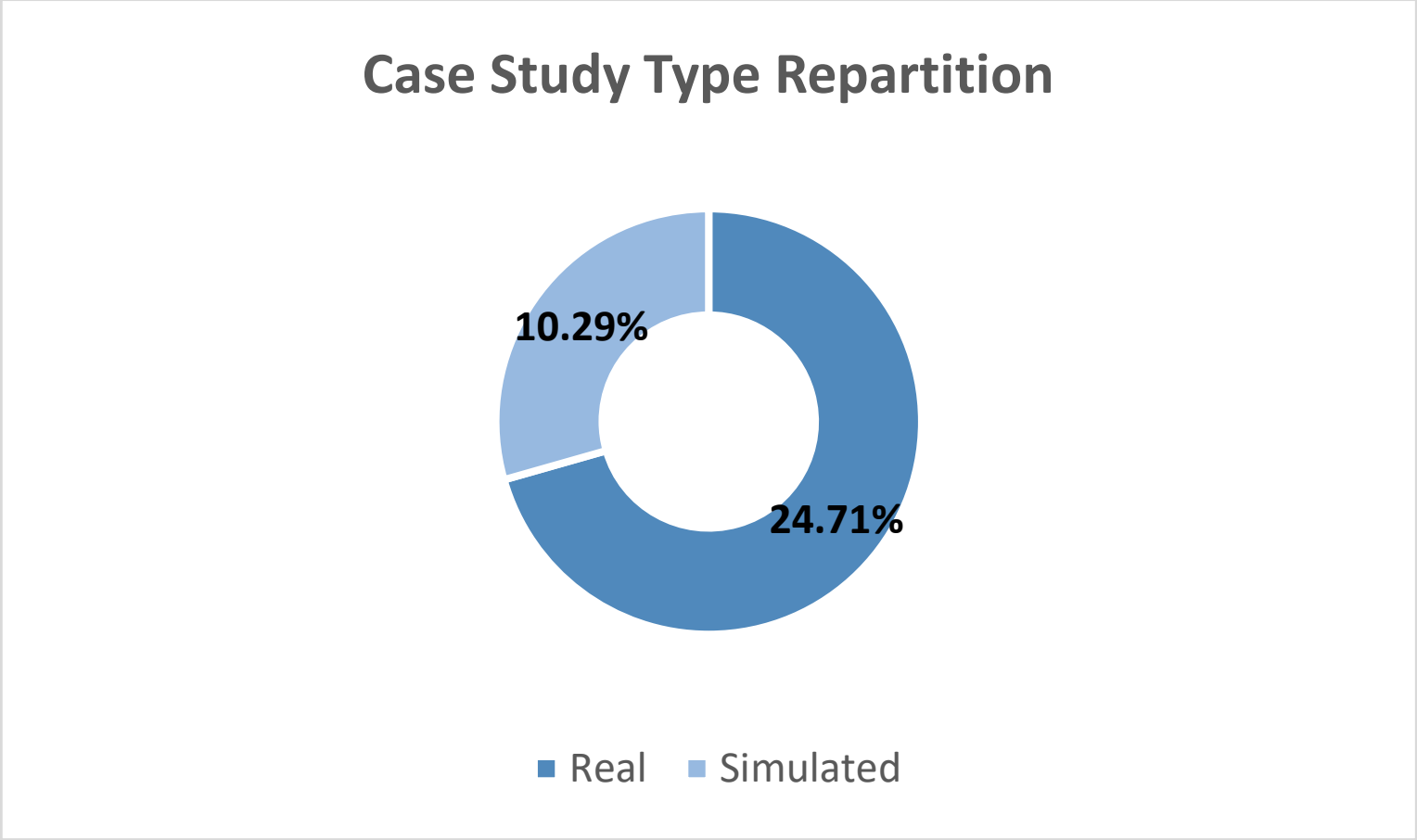

**Figure 14.** Type of Case Study in Selected Works

4.6. Study Implications & Limitations

*4.6.1. Implications*

4.a PHM-XAI early years & interest – Much unexplored opportunity is still available for PHM researchers to advance the assimilation of XAI in PHM.

4.b Interpretable models, rule & knowledge-based models and attention mechanism as most widely used techniques – More research involving other approaches could give additional insight to the PHM community in term of performance, ease of use and flexibility of the XAI method.

4.c XAI as PHM tool and instrument of explanation – XAI could be preferred or required within PHM compared to standalone method.

4.d PHM performance uninfluenced by XAI – Boost the confidence of PHM practitioner and end user in AI model's adoption.

4.e Lack in human role, explanation metrics and uncertainty management – Future efforts need to be concentrated in these areas amongst other in the future. Furthermore, the development of evaluation metrics that can cater PHM needs is urgently recommended.

4.f Mostly real case studies were tested – Boost the confidence of PHM practitioner and end user in AI model's adoption.

*4.6.2. Limitations*

4.g This review does not classify XAI methods in term of its nature : post-hoc, local or global explainability – New insight or pattern could potentially be uncovered by applying this categorization.

4.h This review does not assess the various definitions of interpretability and its derivations that exist in PHM-XAI works.

4.i The classification of performance in qualitative form dilutes the information. A quantitative classification permits comparison to be done especially when a common dataset is used —e.g., turbofan simulated dataset. Comparison in term of model's task—i.e., regression, classification, can also be done using quantitative assessment.

## 5. CONCLUSION

In this work, a state-of-the-art systematic review on XAI applications linked to PHM of industrial asset is compiled. The review follows the guidelines of PRISMA for best practice in systematic review reporting. 35 peer reviewed articles, in English language, from 2015 to 2021, on the subject were selected and examined to accomplish the review objectives. Several interesting findings were uncovered. Firstly, this review found that XAI is attracting interest in PHM domain, with a spike of published works in 2020, though still in its infancy phase. Interpretable model, rule & knowledge based as well as attention mechanism are the most widely used XAI techniques applied in PHM works. Secondly, XAI is central to PHM, assimilated as a tool to execute PHM tasks by the majority of diagnostic and anomaly detection works, while simultaneously being an instrument of explanation. Thirdly, it is discovered that PHM performance is unaltered by XAI. In fact, the majority of PHM-XAI works achieved excellent performance while the rest produce only good results. There is however much work to be done in term of human participation, explanation metrics and uncertainty management which are nearly absent. Finally, this review discovered that mostly real, industrial case studies are tested to demonstrate the effectiveness of XAI, signifying the readiness of AI and XAI to solve real, complex industrial challenges.

**Table 1.** Analysis Results of Selected Articles

| No | Authors & Year | Title | Publisher & Publication Name | PHM Activity | XAI Approach | Performance | XAI Assist PHM | Metric | Human Role | Uncertainty Management | Case Study |
|---|---|---|---|---|---|---|---|---|---|---|---|
| 1 | [49] Wong et al., 2015 | On equivalence of FIS and ELM for interpretable rule-based knowledge representation | IEEE, IEEE Transactions On Neural Networks And Learning Systems | Diagnostic | Rule & Knowledge-based | Good | Yes | No | No | No | Real – Circulating water system |
| 2 | [51] Wu et al.,2018 | K-PdM: KPI-Oriented Machinery Deterioration Estimation Framework for Predictive Maintenance Using Cluster-Based Hidden Markov Model | IEEE, IEEE Access | Prognostic | Rule & Knowledge-based | Very Good | No | No | No | Probabilistic state transition model | Simulated – Turbofan Engine |
| 3 | [59] Massimo et al., 2018 | Unsupervised classification of multichannel profile data using PCA: an application to an emission control system | Elsevier, Computers & Industrial Engineering | Diagnostic | Cluster-based | Very good | Yes | No | Yes | No | Real – Emission Control System |
| 4 | [40] Mathias et al, 2019 | Forecasting remaining useful life: interpretable deep learning approach via variational bayesian inferences | Elsevier, Decision Support Systems | Prognostic | Interpretable Model | Better than other methods, except LSTM | No | No | No | Uncertainty in model parameters | Simulated – Turbofan Engine |
| 5 | [48] Imene et al., 2019 | Fault isolation in manufacturing systems based on learning algorithm and fuzzy rule selection | Springer, Neural Computing And Applications | Diagnostic | Rule & Knowledge-based | Very good | Yes | No | No | Probabilistic classification by Bayes decision rule | Real – Rotary Kiln |
| 6 | [52] Kerelous et al., 2019 | Interpretable logic tree analysis: A data-driven fault tree methodology for causality analysis | Elsevier, Expert Systems With Applications | Diagnostic | LAD | Very Good | Yes | No | Yes | FTA – Expert opinion | Simulated – Actuator system |
| 7 | [53] Rajendran et al., 2019 | Unsupervised Wireless Spectrum Anomaly Detection With Interpretable Features | IEEE, IEEE Transactions On Cognitive Communications And Networking | Anomaly Detection | Autoencoder | Generally better than other tested methods | Yes | No | No | Probabilistic classification error by discriminator | Real – Software Defined Radio Spectrum<br>Simulated – Synthetic Data |
| 8 | [60] Wang et al., 2019 | An Attention-augmented Deep Architecture for Hard Drive Status Monitoring in Large-scale Storage Systems | ACM, ACM Transactions On Storage | Prognostic, Diagnostic | Attention Mechanism | Generally, better than other methods in prognostic. No comparison in diagnostic | Diagnostic Yes<br>Prognostic No | No | No | No | Real –<br>Hard drive |
| 9 | [69] Le et al., 2019 | Visualization and explainable machine learning for efficient manufacturing and system operations | ASTM, Smart And Sustainable Manufacturing Systems | Diagnostic | Others | N/A[1] | Yes | No | Yes | No | Simulated – Turbofan |
| 10 | [38] Langone et al., 2020 | Interpretable Anomaly Prediction: Predicting anomalous behavior in industry 4.0 settings via regularized logistic regression tools | Elsevier, Data & Knowledge Engineering | Anomaly Detection | Interpretable Model | Good | Yes | No | No | Statistical feature extraction | Real –<br>High pressure plunger pump |
| 11 | [39] Peng et al., 2020 | A dynamic structure-adaptive symbolic approach for slewing bearings' life prediction under variable working conditions | Sage, Structural Health Monitoring | Prognostic | Interpretable Model | Better than previous methods | Yes | No | No | No | Real –<br>Slewing Bearings |
| 12 | [41] Ritto et al., 2020 | Digital twin, physics-based model, and machine learning | Elsevier, Mechanical Systems And Signal Processing | Diagnostic | Interpretable Model | Good | No | No | No | No | Not specified – |

[1] N/A = Item not included in the studied work

| | | | | | | | | | | | |
|---|---|---|---|---|---|---|---|---|---|---|---|
| | | applied to damage detection in structures | | | | | | | | | Spring Mass System |
| 13 | [42] Rea et al., 2020 | Progress toward interpretable machine learning based disruption predictors across tokamaks | Taylor & Francis, Fusion Science And Technology | Diagnostic | Interpretable model | N/A | No | No | No | Physic-based indicator | Real DIII – D and JET Tokamaks |
| 14 | [43] Murari et al., 2020 | Investigating the physics of tokamak global stability with interpretable machine learning tools | MDPI, Applied Sciences | Anomaly Detection | Mathematic equation | Good | No | No | No | No | Type unspecified - Tokamak |
| 15 | [44] Zhou et al., 2020 | Fault diagnosis of gas turbine based on partly interpretable convolutional neural networks | Elsevier, Energy | Diagnostic | Tree-Based | Better than other tested methods | Yes | No | No | No | Simulated– Gas Turbine Model |
| 16 | [46] Zhou et al., 2020 | Addressing Noise and Skewness in Interpretable Health-Condition Assessment by Learning Model Confidence | MDPI, Sensors | Diagnostic | Rule & Knowledge-based | Good | No | No | No | No | Real – Aircraft structure |
| 17 | [47] Jianbo et al., 2020 | Knowledge extraction and insertion to deep belief network for gearbox fault diagnosis | Elsevier, Knowledge-Based Systems | Diagnostic | Rule & Knowledge-based | Very Good | Yes | No | No | No | Real – Gearbox |
| 18 | [50] Conde et al., 2020 | Isotonic boosting classification rules | Springer, Advances In Data Analysis And Classification | Diagnostic | Rule & Knowledge-based | Good and comparable to other methods | Yes | No | No | No | Real – Induction motor |
| 19 | [54] Antonio et al., 2020 | Using an autoencoder in the design of an anomaly detector for smart manufacturing | Elsevier, Pattern Recognition Letters | Anomaly Detection | Autoencoder | Same as the previous best method | Yes | No | No | No | Simulated – Continuous batch washing equipment |
| 20 | [55] Abid et al., 2020 | Robust interpretable deep learning for intelligent fault | IEEE, IEEE Transactions On | Diagnostic | Filter-Based | Better than other tested methods and | Yes | No | No | No | Real – Electrical and |
| | | diagnosis of induction motors | Instrumentation And Measurement | | | previous works | | | | | mechanical motor |
| 21 | [56] Liu et al., 2020 | Tscatnet: an interpretable cross-domain intelligent diagnosis model with antinoise and few-shot learning capability | IEEE, IEEE Transactions On Instrumentation And Measurement | Diagnostic | Filter-based | Better than other tested methods | Yes | No | No | No | Real – Bearing, Drive train |
| 22 | [57] Li et al., 2020 | Waveletkernelnet: an interpretable deep neural network for industrial intelligent diagnosis. | IEEE, IEEE Transactions On Systems, Man, And Cybernetics: Systems | Diagnostic | Filter-based | Better than other tested methods | Yes | No | No | No | Real – Bearing, Drive train |
| 23 | [61] Kim et al., 2020 | An explainable convolutional neural network for fault diagnosis in linear motion guide | IEEE, IEEE Transactions On Industrial Informatics | Diagnostic | Attention Mechanism | Very good | No | No | No | No | Real – Linear motion guide |
| 24 | [62] Chen et al., 2020 | Vibration signals analysis by explainable artificial intelligence (XAI) approach: application on bearing faults diagnosis | IEEE, IEEE Access | Diagnostic | Attention Mechanism | N/A | No | No | No | No | Real – Rolling Bearing |
| 25 | [63] Sun et al., 2020 | Vision-based fault diagnostics using explainable deep learning with class activation maps | IEEE, IEEE Access | Diagnostic | Attention Mechanism | Very Good | No | No | No | No | Real – Base-excited cantilever beam, water pump system |
| 26 | [64] Oh et al., 2020 | VODCA: Verification of diagnosis using CAM-based approach for explainable process monitoring | MDPI, Sensors | Diagnostic | Attention Mechanism | Good | Yes | No | No | True Positive (TP) and True Negative (TN) indicators | Simulated – Ford motor and Real – Sapphire grinding |
| 27 | [65] Sreenath et al., 2020 | Fouling modeling and prediction approach for heat exchangers using deep learning | Elsevier, International Journal Of Heat | Failure Prediction | Model Agnostic | Very Good | No | No | No | No | Simulated – Heat Exchanger Model |
| | | | And Mass Transfer | | | | | | | | |
| 28 | [66] Hong et al., 2020 | Remaining useful life prognosis for turbofan engine using explainable deep neural networks with dimensionality reduction | MDPI, Sensors | Prognostic | Model Agnostic | Very Good | No | No | No | No | Simulated – Turbofan Engine |
| 29 | [68] Grezmak et al., 2020 | Interpretable Convolutional Neural Network through Layer-wise Relevance Propagation for Machine Fault Diagnosis | IEEE, IEEE Sensors Journal | Diagnostic | LRP | Very Good | No | No | No | No | Real – Induction motor |
| 30 | [70] Ming et al., 2020 | ProtoSteer: Steering Deep Sequence Model with Prototypes | IEEE, IEEE Transactions On Visualization And Computer Graphics | Diagnostic | Others | N/A | Yes | No | Yes | No | Real – Vehicle fault log |
| 31 | [72] Chen et al., 2020 | Frequency-temporal-logic-based bearing fault diagnosis and fault interpretation using Bayesian optimization with Bayesian neural networks | Elsevier, Mechanical Systems and Signal Processing | Diagnostic | Others | Better than other tested methods | Yes | No | No | No | Real – Bearings |
| 32 | [45] Steenwinckel et al., 2021 | FLAGS: A methodology for adaptive anomaly detection and root cause analysis on sensor data streams by fusing expert knowledge with machine learning | Elsevier, Future Generation Computer Systems | Anomaly Detection, Diagnostic | Rule & Knowledge-based | Good in anomaly detection<br>No result for diagnostic | Yes, for both | No | Yes | FMEA & FTA – Expert opinion | Real – Train |
| 33 | [58] Zhang et al., 2021 | A New Interpretable Learning Method for Fault Diagnosis of Rolling Bearings | IEEE, IEEE Transactions On Instrumentation And Measurement | Diagnostic | Cluster-based | Very Good | Yes | No | No | No | Real – Rolling bearing |
| 34 | [67] Onchis et al., 2021 | Stable and explainable deep learning damage prediction for prismatic cantilever steel beam | Elsevier, Computers In Industry | Diagnostic | Model Agnostic | Very good | Yes, by LIME only | Stability-fit compensation index (SFC) - Quality indicator of the explanations | No | Yes | Real – Prismatic cantilever steel beam |
| 35 | [71] Ding et al., 2021 | Stationary subspaces-vector autoregressive with exogenous terms methodology for degradation trend estimation of rolling and slewing bearings | Elsevier, Mechanical Systems And Signal Processing | Prognostic | Others | Better than other tested methods and comparable to previous works | Yes | No | No | No | Real – Rolling and slewing bearings |

3

**Table 2.** Excluded Articles According to Year

| No. | Authors, Date | Title | Publisher, Publication Name | Exclusion Reason |
|---|---|---|---|---|
| 1 | [76] Kumar et al., 2016 | Adaptive cluster tendency visualization and anomaly detection for streaming data | ACM, ACM Transactions On Knowledge Discovery From Data | Non PHM-XAI implementation/case study |
| 2 | [77] Bao et al., 2016 | Improved fault detection and diagnosis using sparse global-local preserving projections | Elsevier, Journal Of Process Control | Process monitoring & anomaly detection |
| 3 | [78] Kozjek et al., 2017 | Interpretative identification of the faulty conditions in a cyclic manufacturing process | Elsevier, Journal Of Manufacturing Systems | Process monitoring & diagnosis |
| 4 | [79] Ragab et al., 2017 | Fault diagnosis in industrial chemical processes using interpretable patterns based on logical analysis of data | Elsevier, Expert Systems With Applications | Process monitoring & fault diagnosis |
| 5 | [80] Tang et al., 2018 | Fisher discriminative sparse representation based on dbn for fault diagnosis of complex system | MDPI, Applied Science | Process monitoring & fault diagnosis |
| 6 | [81] Luo et al., 2018 | Knowledge-data-integrated sparse modeling for batch process monitoring | Elsevier, Chemical Engineering Science | Process anomaly detection & diagnosis |
| 7 | [82] Puggini et al., 2018 | An enhanced variable selection and Isolation Forest based methodology for anomaly detection with oes data | Elsevier, Engineering Applications Of Artificial Intelligence | Process anomaly detection & diagnosis |
| 8 | [83] Cheng et al., 2018 | Monitoring influent measurements at water resource recovery facility using data-driven soft sensor approach | IEEE, IEEE Sensors Journal | Process anomaly detection |
| 9 | [84] Zhang et al., 2018 | Weakly correlated profile monitoring based on sparse multi-channel functional principal component analysis | T&F, IISE Transactions | Process monitoring |
| 10 | [85] Luo et al., 2018 | Industrial process monitoring based on knowledge-data integrated sparse model and two-level deviation magnitude plots | ACS, Industrial & Engineering Chemistry Research | Process monitoring, anomaly detection & diagnosis |
| 11 | [86] Vojíř et al., 2018 | EasyMiner.eu: web framework for interpretable machine learning based on rules and frequent itemsets | Elsevier, Knowledge-Based Systems | Only development version offers anomaly detection |
| 12 | [87] Du et al., 2019 | A condition change detection method for solar conversion efficiency in solar cell manufacturing processes | IEEE, IEEE Transactions On Semiconductor Manufacturing | Process monitoring & anomaly detection |
| 13 | [88] Keneniet et al., 2019 | Evolving rule-based explainable artificial intelligence for unmanned aerial vehicles | IEEE, IEEE Access | Interpret why agent (UAV) deviate from its mission, not because of system failure |
| 14 | [89] Wang et al., 2019 | Dynamic soft sensor development based on convolutional neural networks | ACS, Industrial & Engineering Chemistry Research | Process modelling |
| 15 | [90] Wang et al., 2019 | Explicit and interpretable nonlinear soft sensor models for influent surveillance at a full-scale wastewater treatment plant | Elsevier, Journal Of Process Control | Process monitoring & variable prediction |
| 16 | [91] Liu et al., 2019 | Intelligent online catastrophe assessment and preventive control via a stacked denoising autoencoder | Elsevier, Neurocomputing | Black Box |
| 17 | [92] Bukhsh et al., 2019 | Predictive maintenance using tree-based classification techniques: a case of railway switches | Elsevier, Transportation Research Part C | Predict maintenance need, activity type and maintenance trigger status |
| 18 | [93] Ragab et al., 2019 | Deep understanding in industrial processes by complementing human expertise with interpretable patterns of machine learning | Elsevier, Expert Systems With Applications | Process monitoring & fault diagnosis |
| 19 | [94] Luo et al., 2019 | Sparse robust principal component analysis with applications to fault detection and diagnosis | ACS, Industrial & Engineering Chemistry Research | Process monitoring, fault detection & diagnosis |
| 20 | [95] Jie et al., 2020 | Process abnormity identification by fuzzy logic rules and expert estimated thresholds derived certainty factor | Elsevier, Chemometrics And Intelligent Laboratory Systems | Process anomaly diagnosis |
| 21 | [96] Sajedi et al., 2020 | Dual bayesian inference for risk-informed vibration-based diagnosis | Wiley, Computer-Aided Civil And Infrastructure Engineering | Uncertainty interpretation, not model's interpretation |
| 22 | [97] Sun et al., 2020 | ALVEN: algebraic learning via elastic net for static and dynamic nonlinear model identification | Elsevier, Computers & Chemical Engineering | Process monitoring & variable prediction |
| 23 | [98] Henriques et al., 2020 | Combining k-means and xgboost models for anomaly detection using log datasets | MDPI, Electronics | Anomaly in project, not engineered system |
| 24 | [99] Gorzałczany et al., 2020 | A modern data-mining approach based on genetically optimized fuzzy systems for interpretable and accurate smart-grid stability prediction | MDPI, Energies | Electrical grid demand stability in financial perspective |
| 25 | [100] Müller et al., 2020 | Data or interpretations impacts of information presentation strategies on diagnostic processes | Wiley, Human Factors And Ergonomics In Manufacturing & Service Industries | Experiment with operator effectivity following quality of interpretability |
| 26 | [101] Shriram et al., 2020 | Least squares sparse principal component analysis and parallel coordinates for real-time process monitoring | ACS, Industrial & Engineering Chemistry Research | Process monitoring & diagnosis |
| 27 | [102] Alshraideh et al., 2020 | Process control via random forest classification of profile signals: an application to a tapping process | Elsevier, Journal Of Manufacturing Processes | Process monitoring & anomaly detection |
| 28 | [103] Minghua et al., 2020 | Diagnosing root causes of intermittent slow queries in cloud databases | ACM, Proceedings Of The VLDB Endowment | Diagnosing slow query due to lack of resources, not failure |
| 29 | [104] Shaha et al., 2020 | Performance prediction and interpretation of a refuse plastic fuel fired boiler | IEEE, IEEE Access | Performance prediction |
| 30 | [105] Kovalev et al., 2020 | SurvLIME: a method for explaining machine learning survival models | Elsevier, Knowledge-Based Systems | Medical survival model |
| 31 | [106] Kovalev et al., 2020 | A robust algorithm for explaining unreliable machine learning survival models using the kolmogorov–smirnov bounds | Elsevier, Neural Networks | Medical survival model |
| 32 | [105] Karn et al., 2021 | Cryptomining detection in container clouds using system calls and explainable machine learning | IEEE, IEEE Transactions On Parallel And Distributed Systems | Network attack |
| 33 | [106] Gyula et al., 2021 | Decision trees for informative process alarm definition and alarm-based fault classification | Elsevier, Process Safety And Environmental Protection | Process monitoring & anomaly detection |
| 34 | [107] Zaman et al., 2021 | Fuzzy heuristics and decision tree for classification of statistical feature-based control chart patterns | MDPI, Symmetry | Process monitoring & diagnosis |
| 35 | [108] Li et al., 2021 | DTDR–ALSTM: extracting dynamic time-delays to reconstruct multivariate data for improving attention-based lstm industrial time series prediction models | Elsevier, Knowledge-Based Systems | Process monitoring & variable prediction |

## *Appendix A*

**Table 3.** Search Strategy

| Database & Date | Extracted Paper Qty | Search Field & Keywords | Filters Applied |
|---|---|---|---|
| IEEE Xplore 18/02/21 | 144 | **Using *'Document Title'*:**<br>1. **Document Title:** explainable OR **Document Title:** interpretable, **Search within results:** diagnostic<br>2. **Document Title:** explainable OR **Document Title:** interpretable, **Search within results:** prognostic<br>3. **Document Title:** explainable OR **Document Title:** interpretable, **Search within results:** diagnosis<br>4. **Document Title:** explainable OR **Document Title:** interpretable, **Search within results:** prognosis | Journals, Early Access Article,<br>**Specify Year Range:** |
| | | 5. Document Title: explainable OR **Document Title:** interpretable, **Search within results:** anomaly detection<br>6. **Document Title:** explainable OR **Document Title:** interpretable, **Search within results:** RUL<br>7. **Document Title:** explainable OR **Document Title:** interpretable, **Search within results:** remaining useful life<br>8. **Document Title:** explainable AI OR **Document Title:** explainable machine learning OR **Document Title:** explainable deep learning OR **Document Title**: XAI**, Search within results:** prognostic<br>9. **Document Title:** explainable AI OR **Document Title:** explainable machine learning OR **Document Title:** explainable deep learning OR **Document Title:** XAI**, Search within results:** diagnostic<br>10. **Document Title:** explainable AI OR **Document Title:** explainable machine learning OR **Document Title:** explainable deep learning OR **Document Title:** XAI**, Search within results:** diagnosis<br>11. **Document Title:** explainable AI OR **Document Title:** explainable machine learning OR **Document Title:** explainable deep learning OR **Document Title:** XAI**, Search within results:** prognosis<br>12. **Document Title:** explainable AI OR **Document Title:** explainable machine learning OR **Document Title:** explainable deep learning OR **Document Title:** XAI**, Search within results:** anomaly detection<br>13. **Document Title:** explainable AI OR **Document Title:** explainable machine learning OR **Document Title:** explainable deep learning OR **Document Title:** XAI**, Search within results:** RUL<br>14. **Document Title:** explainable AI OR **Document Title:** explainable machine learning OR **Document Title:** explainable deep learning OR **Document Title:** XAI**, Search within results:** remaining useful life<br>15. **Document Title:** interpretable AI OR **Document Title:** interpretable machine learning OR **Document Title:** interpretable deep learning OR **Document Title:** XAI**, Search within results:** diagnostic<br>16. **Document Title:** interpretable AI OR **Document Title:** interpretable machine learning OR **Document Title:** interpretable deep learning OR **Document Title:** XAI**, Search within results:** prognostic<br>17. **Document Title:** interpretable AI OR **Document Title:** interpretable machine learning) OR **Document Title:** interpretable deep learning OR **Document Title:** XAI**, Search within results:** prognosis<br>18. **Document Title:** interpretable AI OR **Document Title:** interpretable machine learning) OR **Document Title:** interpretable deep learning OR **Document Title:** XAI**, Search within results:** diagnosis<br>19. **Document Title:** interpretable AI OR **Document Title:** interpretable machine learning) OR **Document Title:** interpretable deep learning OR **Document Title:** XAI**, Search within results:** anomaly detection<br>20. **Document Title:** interpretable AI OR **Document Title:** interpretable machine learning) OR **Document Title:** interpretable deep learning OR **Document Title:** XAI**, Search within results:** RUL<br>21. **Document Title:** interpretable AI OR **Document Title:** interpretable machine learning) OR **Document Title:** interpretable deep learning OR **Document Title:** XAI**, Search within results:** remaining useful life<br><br>**Using *‘Abstract’*:**<br>22. **Abstract:** explainable AI OR **Abstract:** explainable machine learning OR **Abstract:** explainable deep learning OR **Abstract:** XAI **, Search within results:** prognostic<br>23. **Abstract:** explainable AI OR **Abstract:** explainable machine learning OR **Abstract:** explainable deep learning OR **Abstract:** XAI **, Search within results:** diagnostic<br>24. **Abstract:** explainable AI OR **Abstract:** explainable machine learning OR **Abstract:** explainable deep learning OR **Abstract:** XAI **, Search within results:** diagnosis<br>25. **Abstract:** explainable AI OR **Abstract:** explainable machine learning OR **Abstract:** explainable deep learning OR **Abstract:** XAI **, Search within results:** prognosis<br>26. **Abstract:** explainable AI OR **Abstract:** explainable machine learning OR **Abstract:** explainable deep learning OR **Abstract:** XAI **, Search within results:** anomaly detection | 2015-2021 |
| | | 27. **Abstract:** explainable AI OR **Abstract:** explainable machine learning OR **Abstract:** explainable deep learning OR **Abstract:** XAI **, Search within results:** RUL<br>28. **Abstract:** explainable AI OR **Abstract:** explainable machine learning OR **Abstract:** explainable deep learning OR **Abstract:** XAI **, Search within results:** remaining useful life<br>29. **Abstract:** interpretable AI OR **Abstract:** interpretable machine learning) OR **Abstract:** interpretable deep learning OR **Abstract:** XAI, **Search within results:** prognostic<br>30. **Abstract:** interpretable AI OR **Abstract:** interpretable machine learning) OR **Abstract:** interpretable deep learning OR **Abstract:** XAI, **Search within results:** diagnostic<br>31. **Abstract:** interpretable AI OR **Abstract:** interpretable machine learning) OR **Abstract:** interpretable deep learning OR **Abstract:** XAI, **Search within results:** prognosis<br>32. **Abstract:** interpretable AI OR **Abstract:** interpretable machine learning) OR **Abstract:** interpretable deep learning OR **Abstract:** XAI, **Search within results:** diagnosis<br>33. **Abstract:** interpretable AI OR **Abstract:** interpretable machine learning) OR **Abstract:** interpretable deep learning OR **Abstract:** XAI, **Search within results:** anomaly detection<br>34. **Abstract:** interpretable AI OR **Abstract:** interpretable machine learning) OR **Abstract:** interpretable deep learning OR **Abstract:** XAI, **Search within results:** RUL<br>35. **Abstract:** interpretable AI OR **Abstract:** interpretable machine learning) OR **Abstract:** interpretable deep learning OR **Abstract:** XAI, **Search within results:** remaining useful life | |
| Science Direct<br>17/02/21 | 607 | **Using *'Title, abstract or author-specified keywords'*:**<br>36. ("explainable" OR "interpretable") AND ("prognostic" OR "diagnostic" OR "prognosis" OR "diagnosis" OR "anomaly detection" OR "RUL" OR "remaining useful life")<br>37. ("explainable AI" OR "explainable machine learning" OR "explainable deep learning" OR "XAI") AND ("prognostic" OR "diagnostic" OR "anomaly detection" OR "RUL" OR "remaining useful life")<br>38. ("explainable AI" OR "explainable machine learning" OR "explainable deep learning" OR "XAI") AND ("prognosis" OR "diagnosis" OR "anomaly detection" OR "RUL" OR "remaining useful life")<br>39. ("interpretable AI" OR "interpretable machine learning" OR "interpretable deep learning" OR "XAI") AND ("prognostic" OR "diagnostic" OR "anomaly detection" OR "RUL" OR "remaining useful life")<br>40. ("interpretable AI" OR "interpretable machine learning" OR "interpretable deep learning" OR "XAI") AND ("prognosis" OR "diagnosis" OR "anomaly detection" OR "RUL" OR "remaining useful life") | **Article type:** Research Articles,<br>**Subject areas:** Engineering & Computer Science,<br>**Year(s):** 2015-2021 |
| Springer Link<br>22/02/21 | 291 | **Using *'With all the words'*:**<br>41. "explainable" OR "interpretable" AND "prognos"<br>42. "explainable" OR "interpretable" AND "prognos"<br>43. "explainable" OR "interpretable" AND "diagnos"<br>44. "explainable" OR "interpretable" AND "diagnos"<br>45. "explainable" OR "interpretable" AND "RUL"'<br>46. "explainable" OR "interpretable" AND "RUL"<br>47. "explainable" OR "interpretable" AND "remaining useful life"<br>48. "explainable" OR "interpretable" AND "remaining useful life"<br>49. "explainable" OR "interpretable" AND "anomaly detection"<br>50. "explainable" OR "interpretable" AND "anomaly detection" | **Content Type:** Article,<br>**Discipline:** Computer Science or Engineering,<br>**Language:** English,<br>**Show documents** |
| | | | **published:** 2015-2021 |
| ACM Digital Library<br>28/05/21 | 75 | Using ***'Publication Title, Abstract & Keywords'*:**<br>51. **Publication Title:** explainable or interpretable AND **Publication Title:** (prognos OR diagnos OR "anomaly detection" OR RUL OR "remaining useful life"<br>52. **Abstract:** explainable or interpretable AND **Abstract:** (prognos OR diagnos OR "anomaly detection" OR RUL OR "remaining useful life"<br>53. **Keywords:** explainable or interpretable AND **Keywords:** (prognos OR diagnos OR "anomaly detection" OR RUL OR "remaining useful life" | **Publications:** Journal, **Content Type:** Research Article, **Publication Date:** 2015-2021 |
| Scopus<br>27/02/21 | 1931 | 54. ("explainable" OR "interpretable") AND ("prognostic" OR "diagnostic" OR "prognosis" OR "diagnosis" OR "anomaly detection" OR "RUL" OR "remaining useful life" ) | ***Limit to:*** **Document type:** Article, **Publication stage:** Final, **Subject Area:** Engineering & Computer Science, **Language:** English, ***Exclude:*** Medical, **Published from:** 2015-2021 |

***Aknowledgement***

This work is financed by Universiti Teknologi Petronas Foundation.

***Declaration of Competing Interest***

The authors declare that they have no known competing interest in any form that could influence the work

## REFERENCES

1. Peter Stone, Rodney Brooks, Erik Brynjolfsson, Ryan Calo, Oren Etzioni, Greg Hager, Julia Hirschberg, Shivaram Kalyanakrishnan, Ece Kamar, Sarit Kraus, Kevin Leyton-Brown, David Parkes, William Press, AnnaLee Saxenian, Julie Shah, Milind Tambe, and Astro Teller. "Artificial Intelligence and Life in 2030." One Hundred Year Study on Artificial Intelligence: Report of the 2015-2016 Study Panel, Stanford University, Stanford, CA, September 2016. Doc: http://ai100.stanford.edu/2016-report. Accessed: September 6, 2016.
2. Bughin, J., Eric Hazan, S. Ramaswamy, Michael Chui, Tera Allas, Peter Dahlstrom, Nicolaus Henke and Monica Trench. “Artificial intelligence: the next digital frontier?” (2017).
3. The International Telecommunication Union (ITU), (2018), Assessing the Economic Impact of Artificial Intelligence, Artificial Intelligence in Service of Business: Creating a Competitive Advantage, St. Petersburg International Economic Forum 2018
4. Ernst, Ekkehard & Merola, Rossana & Samaan, Daniel. (2018). The economics of artificial intelligence: Implications for the future of work. 10.13140/RG.2.2.29802.57283.
5. Rigla M, García-Sáez G, Pons B, Hernando ME. Artificial Intelligence Methodologies and Their Application to Diabetes. J Diabetes Sci Technol. 2018 Mar;12(2):303-310. doi: 10.1177/1932296817710475. Epub 2017 May 25. PMID: 28539087; PMCID: PMC5851211.
6. Zhang, Xin & Dahu, Wang. (2019). Application of Artificial Intelligence Algorithms in Image Processing. Journal of Visual Communication and Image Representation. 61. 10.1016/j.jvcir.2019.03.004.
7. Xu, Zhaoyi & Saleh, Joseph. (2020). Machine Learning for Reliability Engineering and Safety Applications: Review of Current Status and Future Opportunities.
8. Voulodimos, Athanasios & Doulamis, Nikolaos & Doulamis, Anastasios & Protopapadakis, Eftychios. (2018). Deep Learning for Computer Vision: A Brief Review. Computational Intelligence and Neuroscience. 2018. 1-13. 10.1155/2018/7068349.
9. Linardatos, P.; Papastefanopoulos, V.; Kotsiantis, S. Explainable AI: A Review of Machine Learning Interpretability Methods. Entropy 2021, 23, 18. https://doi.org/10.3390/e23010018
10. S. Zhang, S. Zhang, B. Wang and T. G. Habetler, "Deep Learning Algorithms for Bearing Fault Diagnostics—A Comprehensive Review," in IEEE Access, vol. 8, pp. 29857-29881, 2020, doi: 10.1109/ACCESS.2020.2972859.
11. S. Lu, H. Chai, A. Sahoo and B. T. Phung, "Condition Monitoring Based on Partial Discharge Diagnostics Using Machine Learning Methods: A Comprehensive State-of-the-Art Review," in IEEE Transactions on Dielectrics and Electrical Insulation, vol. 27, no. 6, pp. 1861-1888, December 2020, doi: 10.1109/TDEI.2020.009070.
12. A. L. Ellefsen, V. Æsøy, S. Ushakov and H. Zhang, "A Comprehensive Survey of Prognostics and Health Management Based on Deep Learning for Autonomous Ships," in IEEE Transactions on Reliability, vol. 68, no. 2, pp. 720-740, June 2019, doi: 10.1109/TR.2019.2907402.
13. J. W. Sheppard, M. A. Kaufman and T. J. Wilmer, "IEEE Standards for Prognostics and Health Management," in IEEE Aerospace and Electronic Systems Magazine, vol. 24, no. 9, pp. 34-41, Sept. 2009, doi: 10.1109/MAES.2009.5282287.
14. J. Zhou, L. Zheng, Y. Wang and C. Gogu, "A Multistage Deep Transfer Learning Method for Machinery Fault Diagnostics Across Diverse Working Conditions and Devices," in IEEE Access, vol. 8, pp. 80879-80898, 2020, doi: 10.1109/ACCESS.2020.2990739.
15. Khan, S. and Yairi, T., “A review on the application of deep learning in system health management”, Mechanical Systems and Signal Processing, vol. 107, pp. 241–265, 2018. doi:10.1016/j.ymssp.2017.11.024.
16. G. Aydemir and K. Paynabar, "Image-Based Prognostics Using Deep Learning Approach," in IEEE Transactions on Industrial Informatics, vol. 16, no. 9, pp. 5956-5964, Sept. 2020, doi: 10.1109/TII.2019.2956220.

17. J. J. A. Costello, G. M. West and S. D. J. McArthur, "Machine Learning Model for Event-Based Prognostics in Gas Circulator Condition Monitoring," in IEEE Transactions on Reliability, vol. 66, no. 4, pp. 1048-1057, Dec. 2017, doi: 10.1109/TR.2017.2727489.
18. C. Yang, B. Gunay, Z. Shi and W. Shen, "Machine Learning-Based Prognostics for Central Heating and Cooling Plant Equipment Health Monitoring," in IEEE Transactions on Automation Science and Engineering, vol. 18, no. 1, pp. 346-355, Jan. 2021, doi: 10.1109/TASE.2020.2998586.
19. S. -K. S. Fan, C. -Y. Hsu, D. -M. Tsai, F. He and C. -C. Cheng, "Data-Driven Approach for Fault Detection and Diagnostic in Semiconductor Manufacturing," in IEEE Transactions on Automation Science and Engineering, vol. 17, no. 4, pp. 1925-1936, Oct. 2020, doi: 10.1109/TASE.2020.2983061.
20. X. Li, W. Zhang, N. Xu and Q. Ding, "Deep Learning-Based Machinery Fault Diagnostics With Domain Adaptation Across Sensors at Different Places," in IEEE Transactions on Industrial Electronics, vol. 67, no. 8, pp. 6785-6794, Aug. 2020, doi: 10.1109/TIE.2019.2935987.
21. Q. Yang, X. Jia, X. Li, J. Feng, W. Li and J. Lee, "Evaluating Feature Selection and Anomaly Detection Methods of Hard Drive Failure Prediction," in IEEE Transactions on Reliability, vol. 70, no. 2, pp. 749-760, June 2021, doi: 10.1109/TR.2020.2995724.
22. Tosun AB, Pullara F, Becich MJ, Taylor DL, Fine JL, Chennubhotla SC. Explainable AI (xAI) for Anatomic Pathology. Adv Anat Pathol. 2020 Jul;27(4):241-250. doi: 10.1097/PAP.0000000000000264. PMID: 32541594.
23. Taylor, J.E.T., Taylor, G.W. Artificial cognition: How experimental psychology can help generate explainable artificial intelligence. Psychon Bull Rev 28, 454–475 (2021). https://doi.org/10.3758/s13423-020-01825-5
24. Markus, Aniek & Kors, Jan & Rijnbeek, Peter. (2020). The role of explainability in creating trustworthy artificial intelligence for health care: a comprehensive survey of the terminology, design choices, and evaluation strategies.
25. Jiménez-Luna, J., Grisoni, F. & Schneider, G. Drug discovery with explainable artificial intelligence. Nat Mach Intell 2, 573–584 (2020). https://doi.org/10.1038/s42256-020-00236-4
26. Alejandro Barredo Arrieta, Natalia Díaz-Rodríguez, Javier Del Ser, Adrien Bennetot, Siham Tabik, Alberto Barbado, Salvador Garcia, Sergio Gil-Lopez, Daniel Molina, Richard Benjamins, Raja Chatila, Francisco Herrera, Explainable Artificial Intelligence (XAI): Concepts, taxonomies, opportunities and challenges toward responsible AI, Information Fusion, Volume 58, 2020, Pages 82-115, ISSN 1566-2535, https://doi.org/10.1016/j.inffus.2019.12.012.
27. Payrovnaziri SN, Chen Z, Rengifo-Moreno P, Miller T, Bian J, Chen JH, Liu X, He Z. Explainable artificial intelligence models using real-world electronic health record data: a systematic scoping review. J Am Med Inform Assoc. 2020 Jul 1;27(7):1173-1185. doi: 10.1093/jamia/ocaa053. PMID: 32417928; PMCID: PMC7647281.
28. Stepin, Ilia & Alonso, Jose & Catala, Alejandro & Pereira-Farina, Martin. (2021). A Survey of Contrastive and Counterfactual Explanation Generation Methods for Explainable Artificial Intelligence. IEEE Access. 9. 11974-12001. 10.1109/ACCESS.2021.3051315.
29. Bussmann, Niklas & Giudici, Paolo & Marinelli, Dimitri & Papenbrock, Jochen. (2020). Explainable AI in Fintech Risk Management. Frontiers in Artificial Intelligence. 3. 10.3389/frai.2020.00026.
30. Tjoa E, Guan C. A Survey on Explainable Artificial Intelligence (XAI): Toward Medical XAI. IEEE Transactions on Neural Networks and Learning Systems. 2020 Oct;PP. DOI: 10.1109/tnnls.2020.3027314.
31. Streich, Jared & Romero, Jonathon & Gazolla, João & Kainer, David & Cliff, Ashley & Prates, Erica & Brown, James & Khoury, Sacha & Tuskan, Gerald & Garvin, Michael & Jacobson, Dan & Harfouche, Antoine. (2020). Can exascale computing and explainable artificial intelligence applied to plant biology deliver on the United Nations sustainable development goals?. Current Opinion in Biotechnology. 61. 217-225. 10.1016/j.copbio.2020.01.010.

32. Adadi, Amina & Berrada, Mohammed. (2018). Peeking Inside the Black-Box: A Survey on Explainable Artificial Intelligence (XAI). IEEE Access. PP. 1-1. 10.1109/ACCESS.2018.2870052.
33. Chen, Kexin et al. "Neurorobots as a Means Toward Neuroethology and Explainable AI." Frontiers in neurorobotics vol. 14 570308. 19 Oct. 2020, doi:10.3389/fnbot.2020.570308
34. Xu, Feiyu & Uszkoreit, Hans & Du, Yangzhou & Fan, Wei & Zhao, Dongyan & Zhu, Jun. (2019). Explainable AI: A Brief Survey on History, Research Areas, Approaches and Challenges. 10.1007/978-3-030-32236-6_51.
35. Page MJ, McKenzie JE, Bossuyt PM, Boutron I, Hoffmann TC, Mulrow CD, et al. The PRISMA 2020 statement: an updated guideline for reporting systematic reviews. PLOS Medicine 2021;18(3):e1003583. doi: 10.1371/journal.pmed.1003583
36. Xing, Meng & Yan, Xiaoning & Sun, Xiaoying & Wang, Shoumei & Zhou, Mi & Zhu, Bo & Kuai, Le & Liu, Liu & Luo, Ying & Li, Xin & Li, Bin. (2019). Fire needle therapy for moderate-severe acne: A PRISMA systematic review and meta-analysis of randomized controlled trials. Complementary Therapies in Medicine. 44. 10.1016/j.ctim.2019.04.009.
37. Li, Ting & Hua, Fang & Dan, Shiqi & Zhong, Yuxin & Levey, Colin & Song, Yaling. (2020). Reporting quality of systematic review abstracts in operative dentistry: An assessment using the PRISMA for Abstracts guidelines. Journal of dentistry. 102. 103471. 10.1016/j.jdent.2020.103471.
38. Langone, Rocco & Cuzzocrea, Alfredo & Skantzos, Nikolaos. (2020). Interpretable Anomaly Prediction: Predicting anomalous behavior in industry 4.0 settings via regularized logistic regression tools. Data & Knowledge Engineering. 130. 101850. 10.1016/j.datak.2020.101850.
39. Ding, Peng & Jia, Minping & Wang, Hua. (2020). A dynamic structure-adaptive symbolic approach for slewing bearings' life prediction under variable working conditions. Structural Health Monitoring. 20. 10.1177/1475921720929939.
40. Kraus, Mathias & Feuerriegel, Stefan. (2019). Forecasting remaining useful life: Interpretable deep learning approach via variational Bayesian inferences. Decision Support Systems. 125. 113100. 10.1016/j.dss.2019.113100.
41. Ritto, T.G. & Rochinha, Fernando. (2020). Digital twin, physics-based model, and machine learning applied to damage detection in structures.
42. Rea, C & Montes, K & Pau, Alessandro & Granetz, R & Sauter, O.. (2020). Progress Toward Interpretable Machine Learning- Based Disruption Predictors Across Tokamaks Progress Toward Interpretable Machine Learning-Based Disruption Predictors Across Tokamaks. Fusion Science and Technology. 76. 10.1080/15361055.2020.1798589.
43. Murari A, Peluso E, Lungaroni M, Rossi R, Gelfusa M, Contributors J. Investigating the Physics of Tokamak Global Stability with Interpretable Machine Learning Tools. Applied Sciences. 2020; 10(19):6683. https://doi.org/10.3390/app10196683
44. Dengji Zhou, Qinbo Yao, Hang Wu, Shixi Ma, Huisheng Zhang, Fault diagnosis of gas turbine based on partly interpretable convolutional neural networks, Energy, Volume 200, 2020, 117467, ISSN 0360-5442, https://doi.org/10.1016/j.energy.2020.117467.
45. Bram Steenwinckel, Dieter De Paepe, Sander Vanden Hautte, Pieter Heyvaert, Mohamed Bentefrit, Pieter Moens, Anastasia Dimou, Bruno Van Den Bossche, Filip De Turck, Sofie Van Hoecke, Femke Ongenae, FLAGS: A methodology for adaptive anomaly detection and root cause analysis on sensor data streams by fusing expert knowledge with machine learning, Future Generation Computer Systems, Volume 116, 2021, Pages 30-48, ISSN 0167-739X, https://doi.org/10.1016/j.future.2020.10.015.
46. Zhou Y, Hong S, Shang J, Wu M, Wang Q, Li H, Xie J. Addressing Noise and Skewness in Interpretable Health-Condition Assessment by Learning Model Confidence. Sensors. 2020; 20(24):7307. https://doi.org/10.3390/s20247307

47. Yu, Jianbo & Liu, Guoliang. (2020). Knowledge extraction and insertion to deep belief network for gearbox fault diagnosis. Knowledge-Based Systems. 197. 105883. 10.1016/j.knosys.2020.105883.
48. Djelloul, Imene & Sari, Zaki & Souier, Mehdi. (2019). Fault isolation in manufacturing systems based on learning algorithm and fuzzy rule selection. Neural Computing and Applications. 31. 10.1007/s00521-017-3169-3.
49. S. Y. Wong, K. S. Yap, H. J. Yap, S. C. Tan and S. W. Chang, "On Equivalence of FIS and ELM for Interpretable Rule-Based Knowledge Representation," in IEEE Transactions on Neural Networks and Learning Systems, vol. 26, no. 7, pp. 1417-1430, July 2015, doi: 10.1109/TNNLS.2014.2341655.
50. Conde, D., Fernández, M.A., Rueda, C. *et al.* Isotonic boosting classification rules. *Adv Data Anal Classif* **15,** 289–313 (2021). https://doi.org/10.1007/s11634-020-00404-9
51. Z. Wu et al., "K-PdM: KPI-Oriented Machinery Deterioration Estimation Framework for Predictive Maintenance Using Cluster-Based Hidden Markov Model," in IEEE Access, vol. 6, pp. 41676-41687, 2018, doi: 10.1109/ACCESS.2018.2859922.
52. Waghen, Kerelous & Ouali, Mohamed-Salah. (2019). Interpretable Logic Tree Analysis: A Data-Driven Fault Tree Methodology for Causality Analysis. Expert Systems with Applications. 136. 10.1016/j.eswa.2019.06.042.
53. S. Rajendran, W. Meert, V. Lenders and S. Pollin, "Unsupervised Wireless Spectrum Anomaly Detection With Interpretable Features," in IEEE Transactions on Cognitive Communications and Networking, vol. 5, no. 3, pp. 637-647, Sept. 2019, doi: 10.1109/TCCN.2019.2911524.
54. Antonio L. Alfeo, Mario G.C.A. Cimino, Giuseppe Manco, Ettore Ritacco, Gigliola Vaglini, Using an autoencoder in the design of an anomaly detector for smart manufacturing, Pattern Recognition Letters, Volume 136, 2020, Pages 272-278, ISSN 0167-8655, https://doi.org/10.1016/j.patrec.2020.06.008.
55. F. B. Abid, M. Sallem and A. Braham, "Robust Interpretable Deep Learning for Intelligent Fault Diagnosis of Induction Motors," in IEEE Transactions on Instrumentation and Measurement, vol. 69, no. 6, pp. 3506-3515, June 2020, doi: 10.1109/TIM.2019.2932162.
56. C. Liu, C. Qin, X. Shi, Z. Wang, G. Zhang and Y. Han, "TScatNet: An Interpretable Cross-Domain Intelligent Diagnosis Model With Antinoise and Few-Shot Learning Capability," in IEEE Transactions on Instrumentation and Measurement, vol. 70, pp. 1-10, 2021, Art no. 3506110, doi: 10.1109/TIM.2020.3041905.
57. T. Li et al., "WaveletKernelNet: An Interpretable Deep Neural Network for Industrial Intelligent Diagnosis," in IEEE Transactions on Systems, Man, and Cybernetics: Systems, doi: 10.1109/TSMC.2020.3048950.
58. D. Zhang, Y. Chen, F. Guo, H. R. Karimi, H. Dong and Q. Xuan, "A New Interpretable Learning Method for Fault Diagnosis of Rolling Bearings," in IEEE Transactions on Instrumentation and Measurement, vol. 70, pp. 1-10, 2021, Art no. 3507010, doi: 10.1109/TIM.2020.3043873.
59. Pacella, Massimo. (2018). Unsupervised Classification of Multichannel Profile Data using PCA: an application to an Emission Control System. Computers & Industrial Engineering. 122. 10.1016/j.cie.2018.05.029.
60. Ji Wang, Weidong Bao, Lei Zheng, Xiaomin Zhu, and Philip S. Yu. 2019. An Attention-augmented Deep Architecture for Hard Drive Status Monitoring in Large-scale Storage Systems. ACM Trans. Storage 15, 3, Article 21 (August 2019), 26 pages. DOI:https://doi.org/10.1145/3340290
61. M. S. Kim, J. P. Yun and P. Park, "An Explainable Convolutional Neural Network for Fault Diagnosis in Linear Motion Guide," in IEEE Transactions on Industrial Informatics, doi: 10.1109/TII.2020.3012989.

62. H. -Y. Chen and C. -H. Lee, "Vibration Signals Analysis by Explainable Artificial Intelligence (XAI) Approach: Application on Bearing Faults Diagnosis," in IEEE Access, vol. 8, pp. 134246-134256, 2020, doi: 10.1109/ACCESS.2020.3006491.
63. K. H. Sun, H. Huh, B. A. Tama, S. Y. Lee, J. H. Jung and S. Lee, "Vision-Based Fault Diagnostics Using Explainable Deep Learning With Class Activation Maps," in IEEE Access, vol. 8, pp. 129169-129179, 2020, doi: 10.1109/ACCESS.2020.3009852.
64. Oh, C.; Jeong, J. VODCA: Verification of Diagnosis Using CAM-Based Approach for Explainable Process Monitoring. Sensors 2020, 20, 6858. https://doi.org/10.3390/s20236858
65. Sundar, Sreenath & C. Rajagopal, Manjunath & Zhao, Hanyang & Kuntumalla, Gowtham & Meng, Yuquan & Chang, Ho & Shao, Chenhui & Ferreira, Placid & Miljkovic, Nenad & Sinha, Sanjiv & Salapaka, Srinivasa. (2020). Fouling modeling and prediction approach for heat exchangers using deep learning. International Journal of Heat and Mass Transfer. 159. 120112. 10.1016/j.ijheatmasstransfer.2020.120112.
66. Hong CW, Lee C, Lee K, Ko M-S, Kim DE, Hur K. Remaining Useful Life Prognosis for Turbofan Engine Using Explainable Deep Neural Networks with Dimensionality Reduction. Sensors. 2020; 20(22):6626. https://doi.org/10.3390/s20226626
67. Onchis, Darian & Gillich, Gilbert-Rainer. (2021). Stable and explainable deep learning damage prediction for prismatic cantilever steel beam. Computers in Industry. 125. 103359. 10.1016/j.compind.2020.103359.
68. J. Grezmak, J. Zhang, P. Wang, K. A. Loparo and R. X. Gao, "Interpretable Convolutional Neural Network Through Layer-wise Relevance Propagation for Machine Fault Diagnosis," in IEEE Sensors Journal, vol. 20, no. 6, pp. 3172-3181, 15 March15, 2020, doi: 10.1109/JSEN.2019.2958787.
69. Le, Dy & Vung, Pham & Nguyen, Huyen & Dang, Tommy. (2019). Visualization and Explainable Machine Learning for Efficient Manufacturing and System Operations. 3. 20190029. 10.1520/SSMS20190029.
70. Y. Ming, P. Xu, F. Cheng, H. Qu and L. Ren, "ProtoSteer: Steering Deep Sequence Model with Prototypes," in IEEE Transactions on Visualization and Computer Graphics, vol. 26, no. 1, pp. 238-248, Jan. 2020, doi: 10.1109/TVCG.2019.2934267.
71. Peng Ding, Minping Jia, Xiaoan Yan, Stationary subspaces-vector autoregressive with exogenous terms methodology for degradation trend estimation of rolling and slewing bearings, Mechanical Systems and Signal Processing, Volume 150, 2021, 107293, ISSN 0888-3270, https://doi.org/10.1016/j.ymssp.2020.107293.
72. Chen, Gang & Liu, Mei & Chen, Jin. (2020). Frequency-temporal-logic-based bearing fault diagnosis and fault interpretation using Bayesian optimization with Bayesian neural networks. Mechanical Systems and Signal Processing. 145. 106951. 10.1016/j.ymssp.2020.106951.
73. Zhou J, Gandomi AH, Chen F, Holzinger A. Evaluating the Quality of Machine Learning Explanations: A Survey on Methods and Metrics. Electronics. 2021; 10(5):593. https://doi.org/10.3390/electronics10050593
74. Martin, K., Liret, A., Wiratunga, N. et al. Evaluating Explainability Methods Intended for Multiple Stakeholders. Künstl Intell (2021). https://doi.org/10.1007/s13218-020-00702-6
75. Holzinger, A., Carrington, A. & Müller, H. Measuring the Quality of Explanations: The System Causability Scale (SCS). Künstl Intell 34, 193–198 (2020). https://doi.org/10.1007/s13218-020-00636-z

**EXCLUDED ARTICLES**

76. Kumar, Dheeraj & Bezdek, James & Rajasegarar, Sutharshan & Palaniswami, Marimuthu & Leckie, Christopher & Chan, Jeffrey & Gubbi, Jayavardhana. (2016). Adaptive Cluster

Tendency Visualization and Anomaly Detection for Streaming Data. ACM Transactions on Knowledge Discovery from Data. 11. 1-40. 10.1145/2997656.

77. Bao, Shiyi & Luo, Lijia & Mao, Jianfeng. (2016). Improved fault detection and diagnosis using sparse global-local preserving projections. Journal of Process Control. 47. 10.1016/j.jprocont.2016.09.007.
78. Kozjek, Dominik & Vrabič, Rok & Kralj, David & Butala, Peter. (2017). Interpretative identification of the faulty conditions in a cyclic manufacturing process. Journal of Manufacturing Systems. 43. 10.1016/j.jmsy.2017.03.001.
79. Ragab, Ahmed & El Koujok, Mohamed & Poulin, Bruno & Amazouz, Mouloud & Yacout, Soumaya. (2017). Fault Diagnosis in Industrial Chemical Processes Using Interpretable Patterns Based on Logical Analysis of Data. Expert Systems with Applications. 95. 10.1016/j.eswa.2017.11.045.
80. Tang Q, Chai Y, Qu J, Ren H. Fisher Discriminative Sparse Representation Based on DBN for Fault Diagnosis of Complex System. Applied Sciences. 2018; 8(5):795. https://doi.org/10.3390/app8050795
81. Luo, Lijia & Bao, Shiyi. (2018). Knowledge-data-integrated sparse modeling for batch process monitoring. Chemical Engineering Science. 189. 10.1016/j.ces.2018.05.055.
82. Puggini, Luca & Mcloone, Sean. (2018). An enhanced variable selection and Isolation Forest based methodology for anomaly detection with OES data. Engineering Applications of Artificial Intelligence. 67. 126-135. 10.1016/j.engappai.2017.09.021.
83. Cheng, Tuoyuan & Harrou, Fouzi & Sun, Ying & Leiknes, Torove. (2018). Monitoring Influent Measurements at Water Resource Recovery Facility Using Data-Driven Soft Sensor Approach. IEEE Sensors Journal. PP. 1-1. 10.1109/JSEN.2018.2875954.
84. Zhang, Chen & Yan, Hao & Lee, Seungho & Shi, Jianjun. (2018). Weakly Correlated Profile Monitoring Based on Sparse Multi-channel Functional Principal Component Analysis. IISE Transactions. 50. 1-29. 10.1080/24725854.2018.1451012.
85. Industrial Process Monitoring Based on Knowledge–Data Integrated Sparse Model and Two-Level Deviation Magnitude Plots, Lijia Luo, Shiyi Bao, Jianfeng Mao, and Zhenyu Ding, Industrial & Engineering Chemistry Research 2018 57 (2), 611-622, DOI: 10.1021/acs.iecr.7b02150
86. Vojíř, Stanislav & Zeman, Václav & Kuchař, Jaroslav & Kliegr, Tomáš. (2018). EasyMiner.eu: Web Framework for Interpretable Machine Learning based on Rules and Frequent Itemsets. Knowledge-Based Systems. 150. 10.1016/j.knosys.2018.03.006.
87. J. Du, X. Zhang and J. Shi, "A Condition Change Detection Method for Solar Conversion Efficiency in Solar Cell Manufacturing Processes," in IEEE Transactions on Semiconductor Manufacturing, vol. 32, no. 1, pp. 82-92, Feb. 2019, doi: 10.1109/TSM.2018.2875011.
88. B. M. Keneni et al., "Evolving Rule-Based Explainable Artificial Intelligence for Unmanned Aerial Vehicles," in IEEE Access, vol. 7, pp. 17001-17016, 2019, doi: 10.1109/ACCESS.2019.2893141.
89. "Dynamic Soft Sensor Development Based on Convolutional Neural Networks", Kangcheng Wang, Chao Shang, Lei Liu, Yongheng Jiang, Dexian Huang, and Fan Yang, Industrial & Engineering Chemistry Research, 2019 58 (26), 11521-11531 DOI: 10.1021/acs.iecr.9b02513
90. Wang, Xiaodong & Kvaal, Knut & Ratnaweera, Harsha. (2019). Explicit and interpretable nonlinear soft sensor models for influent surveillance at a full-scale wastewater treatment plant. Journal of Process Control. 77. 1-6. 10.1016/j.jprocont.2019.03.005.
91. Liu, Yangyang & Zhai, Mingyu & Jin, Jiahui & Song, Aibo & Lin, Jikeng & Wu, Zhiang & Zhao, Yixin. (2019). Intelligent Online Catastrophe Assessment and Preventive Control via a Stacked Denoising Autoencoder. Neurocomputing. 380. 10.1016/j.neucom.2019.10.090.
92. Bukhsh, Zaharah & Saeed, Aaqib & Stipanovic, Irina & Dorée, André. (2019). Predictive maintenance using tree-based classification techniques: A case of railway switches.

Transportation Research Part C Emerging Technologies. 101. 35-54. 10.1016/j.trc.2019.02.001.

93. Ragab, Ahmed & El Koujok, Mohamed & Ghezzaz, Hakim & Amazouz, Mouloud & Ouali, Mohamed-Salah & Yacout, Soumaya. (2019). Deep Understanding in Industrial Processes by Complementing Human Expertise with Interpretable Patterns of Machine Learning. Expert Systems with Applications. 10.1016/j.eswa.2019.01.011.

94. "Sparse Robust Principal Component Analysis with Applications to Fault Detection and Diagnosis", Lijia Luo, Shiyi Bao, and Chudong Tong, Industrial & Engineering Chemistry Research 2019 58 (3), 1300-1309, DOI: 10.1021/acs.iecr.8b04655"

95. jie, Yuan & shumei, Zhang & shu, Wang & fuli, Wang & luping, Zhao. (2020). Process abnormity identification by fuzzy logic rules and expert estimated thresholds derived certainty factor. Chemometrics and Intelligent Laboratory Systems. 209. 104232. 10.1016/j.chemolab.2020.104232.

96. Sajedi, S, Liang, X. Dual Bayesian inference for risk-informed vibration-based damage diagnosis. Comput Aided Civ Inf. 2020; 1– 17. https://doi.org/10.1111/mice.12642

97. Sun, Weike & Braatz, Richard. (2020). ALVEN: Algebraic learning via elastic net for static and dynamic nonlinear model identification. Computers & Chemical Engineering. 143. 107103. 10.1016/j.compchemeng.2020.107103.

98. Henriques J, Caldeira F, Cruz T, Simões P. Combining K-Means and XGBoost Models for Anomaly Detection Using Log Datasets. Electronics. 2020; 9(7):1164. https://doi.org/10.3390/electronics9071164

99. Gorzałczany MB, Piekoszewski J, Rudziński F. A Modern Data-Mining Approach Based on Genetically Optimized Fuzzy Systems for Interpretable and Accurate Smart-Grid Stability Prediction. Energies. 2020; 13(10):2559. https://doi.org/10.3390/en13102559

100. Müller, R, Gögel, C, Bönsel, R. Data or interpretations: Impacts of information presentation strategies on diagnostic processes. Hum Factors Man. 2020; 30: 266– 281. https://doi.org/10.1002/hfm.20838

101. Least squares sparse principal component analysis and parallel coordinates for real-time process monitoring, Shriram Gajjar, Murat Kulahci, and Ahmet Palazoglu, Industrial & Engineering Chemistry Research 2020 59 (35), 15656-15670, DOI: 10.1021/acs.iecr.0c01749

102. Alshraideh, Hussam & del Castillo, Enrique & Gil Del Val, Alain. (2020). Process control via random forest classification of profile signals: An application to a tapping process. Journal of Manufacturing Processes. 58. 736-748. 10.1016/j.jmapro.2020.08.043.

103. Minghua Ma, Zheng Yin, Shenglin Zhang, Sheng Wang, Christopher Zheng, Xinhao Jiang, Hanwen Hu, Cheng Luo, Yilin Li, Nengjun Qiu, Feifei Li, Changcheng Chen, and Dan Pei. 2020. Diagnosing root causes of intermittent slow queries in cloud databases. Proc. VLDB Endow. 13, 8 (April 2020), 1176–1189. DOI:https://doi.org/10.14778/3389133.3389136

104. P. Shaha, M. S. Singamsetti, B. K. Tripathy, G. Srivastava, M. Bilal and L. Nkenyereye, "Performance Prediction and Interpretation of a Refuse Plastic Fuel Fired Boiler," in IEEE Access, vol. 8, pp. 117467-117482, 2020, doi: 10.1109/ACCESS.2020.3004156.

105. Maxim S. Kovalev, Lev V. Utkin, Ernest M. Kasimov, SurvLIME: A method for explaining machine learning survival models, Knowledge-Based Systems, Volume 203, 2020, 106164, ISSN 0950-7051, https://doi.org/10.1016/j.knosys.2020.106164."

106. Kovalev, Maxim & Utkin, Lev. (2020). A robust algorithm for explaining unreliable machine learning survival models using the Kolmogorov-Smirnov bounds.

107. R. R. Karn, P. Kudva, H. Huang, S. Suneja and I. M. Elfadel, "Cryptomining Detection in Container Clouds Using System Calls and Explainable Machine Learning," in IEEE Transactions on Parallel and Distributed Systems, vol. 32, no. 3, pp. 674-691, 1 March 2021, doi: 10.1109/TPDS.2020.3029088.

108. "Gyula Dorgo, Ahmet Palazoglu, Janos Abonyi, Decision trees for informative process alarm definition and alarm-based fault classification, Process Safety and Environmental Protection,

Volume 149, 2021, Pages 312-324, ISSN 0957-5820, https://doi.org/10.1016/j.psep.2020.10.024."

109. Zaman, Munawar & Hassan, Adnan. (2021). Fuzzy Heuristics and Decision Tree for Classification of Statistical Feature-Based Control Chart Patterns. Symmetry. 13. 110. 10.3390/sym13010110.

110. Li, Jince & Yang, Bo & Li, Hongguang & Wang, Yongjian & Qi, Chu & Liu, Yi. (2021). DTDR–ALSTM: Extracting dynamic time-delays to reconstruct multivariate data for improving attention-based LSTM industrial time series prediction models. Knowledge-Based Systems. 211. 106508. 10.1016/j.knosys.2020.106508.